\documentclass[journal]{IEEEtran} % Comment this line out if you need a4paper

\usepackage{graphics} % for pdf, bitmapped graphics files
\usepackage{graphicx}
\usepackage[dvipsnames]{xcolor,colortbl}
\usepackage{multirow}
\usepackage{amssymb}
\usepackage{multirow, tabularx}
\usepackage{comment}
\usepackage{url}
\usepackage{subcaption}
\usepackage{hyperref}

\newcommand{\etal}{\textit{et al}.~}
\newcommand{\ie}{i.e.,~}
\newcommand{\eg}{e.g.,~}

\begin{document}

\title{\LARGE \bf
Show and Grasp: Few-shot Semantic Segmentation for Robot Grasping through Zero-shot Foundation Models
}

\author{Leonardo Barcellona,~\IEEEmembership{Student Member,~IEEE,} Alberto Bacchin,~\IEEEmembership{Student Member,~IEEE,} Matteo Terreran,~\IEEEmembership{Member,~IEEE,} Emanuele Menegatti,~\IEEEmembership{Senior Member,~IEEE,} Stefano Ghidoni~\IEEEmembership{Member,~IEEE}%
\thanks{L.B., A.B. (Corresponding Author), M.T., E.M. and S.G. are with the Intelligent Autonomous System Lab, Department of Information Engineering, University of Padova, Padua, Italy (e-mail: leonardo.barcellona@phd.unipd.it, bacchinalb@dei.unipd.it, matteo.terreran@dei.unipd.it, emg@dei.unipd.it, stefano.ghidoni@dei.unipd.it).}%
\thanks{L.B. is also with the Politecnico di Torino, 10138 Torino, Italy (email: leonardo.barcellona@polito.it).}%
\thanks{L.B. and A.B. contributed equally to
this work.}%
\thanks{Part of this work was supported by MUR (Italian Minister for University and Research) under the initiative “PON Ricerca e Innovazione 2014 - 2020”, CUP C95F21007870007.}%
%\thanks{Manuscript received XXXX; revised XXXX.}
}

% \markboth{IEEE TRANSACTIONS ON ROBOTICS,~Vol.~XX, No.~X, XXX~2024}%
% {Shell \MakeLowercase{\textit{et al.}}: Bare Demo of IEEEtran.cls for IEEE Journals}

\maketitle
% \thispagestyle{empty}
% \pagestyle{empty}

%%%%%%%%%%%%%%%%%%%%%%%%%%%%%%%%%%%%%%%%%%%%%%%%%%%%%%%%%%%%%%%%%%%%%%%%%%%%%%%%
\begin{abstract}

The ability of a robot to pick an object, known as robot grasping, is crucial for several applications, such as assembly or sorting. In such tasks, selecting the right target to pick is as essential as inferring a correct configuration of the gripper. A common solution to this problem relies on semantic segmentation models, which often show poor generalization to unseen objects and require considerable time and massive data to be trained. To reduce the need for large datasets, some grasping pipelines exploit few-shot semantic segmentation models, which are capable of recognizing new classes given a few examples. However, this often comes at the cost of limited performance and fine-tuning is required to be effective in robot grasping scenarios.
In this work, we propose to overcome all these limitations by combining the impressive generalization capability reached by foundation models with a high-performing few-shot classifier, working as a score function to select the segmentation that is closer to the support set.
%To achieve such behaviour, we propose an inversion of the roles of the classifier's input.
The proposed model is designed to be embedded in a grasp synthesis pipeline. The extensive experiments using one or five examples show that our novel approach overcomes existing performance limitations, improving the state of the art both in few-shot semantic segmentation on the Graspnet-1B (+10.5\% mIoU) and Ocid-grasp (+1.6\% AP) datasets, and real-world few-shot grasp synthesis (+21.7\% grasp accuracy). The project page is available at: \url{https://leobarcellona.github.io/showandgrasp.github.io/}

%\textcolor{blue}{TODO: add repo link after tidy up}
\end{abstract}


%%%%%%%%%%%%%%%%%%%%%%%%%%%%%%%%%%%%%%%%%%%%%%%%%%%%%%%%%%%%%%%%%%%%%%%%%%%%%%%%
\section{INTRODUCTION}
\label{sec:introduction}
% Togliere menzione a SAM e spingere zero shot, poi aggiungere motivazione test di cutler
% Contibuto è abbiamo adattato un'architettura di grasp synthesis per poter integrare il few-shot seg 
% di non dire backbone rimossa dal nostro paper icra, ma diciamo che noi abbiamo fatto diverso perchè segmentazione è più precisa
% dire che nel real-word proponiamo anche due esprimenti che dimostrano che l'approccio funziona su per selezionare l'oggetto o usare un supporto diverso. 

Grasp synthesis is the task of determining which is the best pose of a gripper in order to pick an object.
%As reported in~\cite{review_grasp_tro}, 
This is an essential aspect of robotics, as it is required for many manipulation tasks~\cite{seita2021learningMANIPULATION, shridhar2023perceiverMANIPULATION2, geng2023rlaffordMANIPULATION3} that enable robots to interact with the environment.
In simple cases, grasp poses can be inferred from geometrical information that are extracted from depth images or point clouds~\cite{ggcnn}.
%This is done by inferring a grasp pose or a set of grasp poses~\cite{review_grasp_tro}, usually based on geometrical information that can be extracted from depth images or point clouds~\cite{ggcnn}.
%often based on the typical sensory information available in such contexts when dealing with individual objects, namely images or point clouds~\cite{ggcnn}. 
However, more complex scenarios also exist: when a robot operates in cluttered environments with multiple objects, the selection of the target to be grasped should be performed in addition to tackling the geometric problem of generating a suitable grasp pose.
Target selection can be done by generating multiple grasps for each object, and then selecting the grasp with the highest success rate. However, this shall be harmonized with other constraints, including semantic information of the objects in the scene and the choice of the object to be grasped, which is relevant in a number of applications like assembly \cite{assembly}, sorting \cite{sorting} or object handover \cite{christen2023learningHANDOVER}. 

The use of semantic information within grasp generation is gaining great attention in the research community~\cite{duan2023semanticGRASPSEMANTIC}.
%-------------This work falls in the field, as we incorporate semantic information into the grasp generation process, as it can help to deduce optimal grasp poses for a specific target object.
In several works, the selection of the best grasp pose of a given target exploits information coming from the semantic segmentation of the scene~\cite{duan2023semanticGRASPSEMANTIC,ainetter2021endOCIDGRASP}, based on Deep Learning (DL) models trained on a specific set of classes.
%
%%%%%%%modifica proposts da EM
%Several works exploit the semantic segmentation of the scene to select the best grasp pose for a given target. The semantic segmentation is obtained with Deep Learning (DL) models trained on a specific set of classes ~\cite{duan2023semanticGRASPSEMANTIC,ainetter2021endOCIDGRASP}. 

\begin{figure}[t!]
\centering
\resizebox{0.99\columnwidth}{!}{
%\framebox{\parbox{3in}{Add image showing skeletons VS 3D Segmentation for qualitative comparison}}
\includegraphics[]{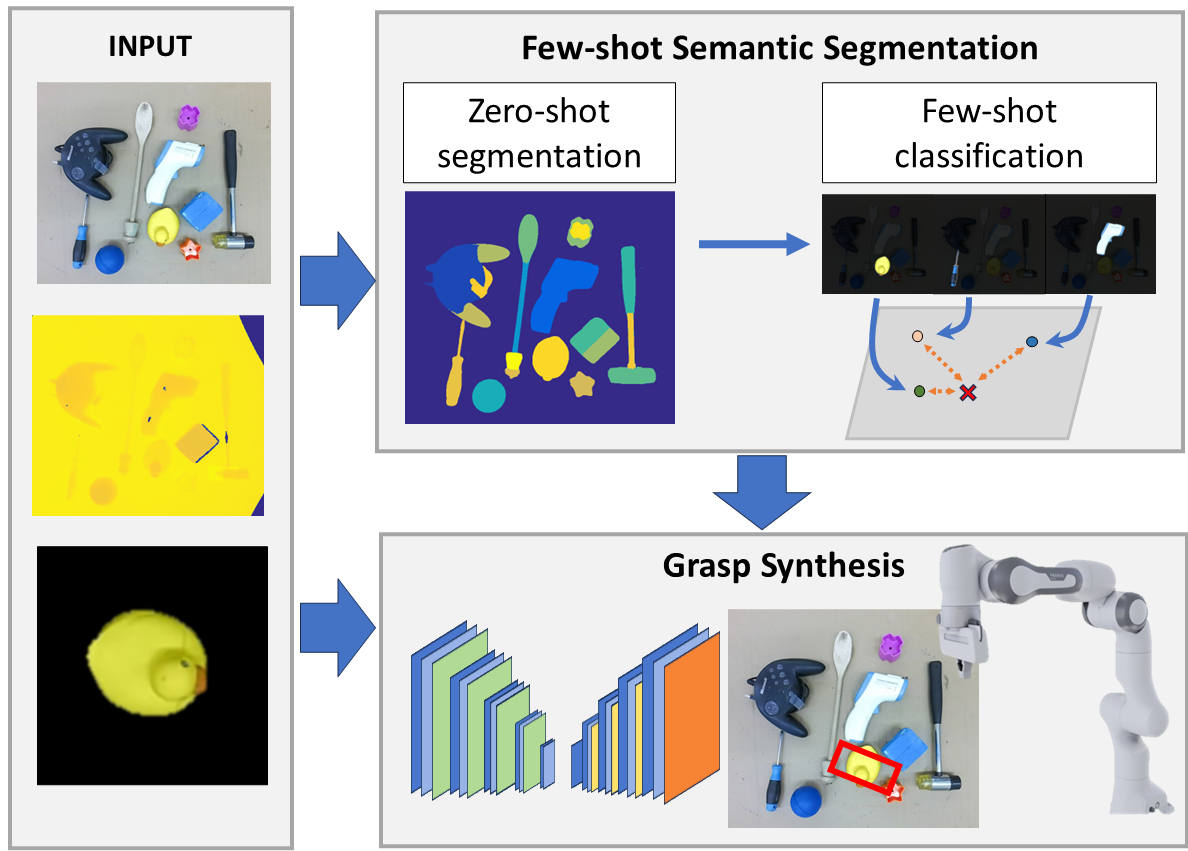}
}
\caption{Representation of the pipeline. The input is an RGB-D image and an example of the target object. To select the object, the segmentation is computed using a zero-shot segmentator and a few-shot classifier. The produced segmentation and the depth image are used to infer the grasp pose.}
\label{fig:approach}
\end{figure}
However, these approaches have limitations that arise from the underlying supervised DL models, including the need for a large number of labelled examples and the capability of recognizing only the object classes observed at training time. These constraints limit the applicability in several real-world scenarios where new object classes are commonly encountered, and it is time-consuming or difficult to collect large datasets for such classes.
 
To address these limitations, Few-shot Semantic Grasping (FSG) was recently introduced~\cite{liu2022unseenFSSRG, holomjova2023gsmrGSMR, barcellona2023fsgFSGNET}. The objective of FSG is to generate suitable grasping poses for an unseen object in a cluttered scene, using a small set of images (called support) by synthesizing the grasp by means of Few-Shot Segmentation (FSS) models, that can manage new classes given only a very limited number of labelled examples (shots) at inference time. This work aims to advance Few-shot Semantic Grasping by enhancing the generalization capabilities of FSS.
Although FSS models have good generalization, they exhibit a significant drop in performance when the test dataset is sensibly different from the training data~\cite{boudiaf2021fewNOMETA}, which restricts the applicability of such systems.
%This is a major shortcoming as it restricts the applicability of such systems to those scenarios that are similar to the one observed at training time---
This effect can be mitigated thanks to fine-tuning which, however, does not ensure the generation of high-quality segmentation masks for robotic grasping ~\cite{barcellona2023fsgFSGNET}.
While FSS struggles to generalize to novel scenarios, the similar task of Few-Shot Classification (FSC), whose objective is labelling an image among a set of new classes not used for the training, was able to reach impressive results~\cite{hu2022pushingPMF}, taking advantage of the introduction of Vision Transformers (ViT)~\cite{dosovitskiy2020imageVIT}. 
%
%Further, the dawn of foundation models in computer vision is changing how models are designed \cite{wei2023iclipICLIP, kirillov2023segmentSAM, wang2023cutCUTLER}. Thanks to their impressive zero-shot generalization, researchers directly build their solutions upon them, often without even fine-tuning their knowledge \cite{ye2023gaussian3DRECONSTRUCTION, qin2023langsplatRECONSTRUCTIONLANGUAGE}.   
%
%In this work, we propose a novel approach for few-shot semantic segmentation based zero-shot segmentation from foundation models\cite{kirillov2023segmentSAM, wang2023cutCUTLER} and an high performing few-shot classifier~\cite{hu2022pushingPMF}. 
%On the other hand, the dawn of foundation models in computer vision~\cite{wei2023iclipICLIP, kirillov2023segmentSAM, wang2023cutCUTLER} is leading to models which do not require fine-tuning, owing to their impressive zero-shot generalizability~\cite{ye2023gaussian3DRECONSTRUCTION, qin2023langsplatRECONSTRUCTIONLANGUAGE}. 
The advent of foundation models in computer vision~\cite{wei2023iclipICLIP, kirillov2023segmentSAM, wang2023cutCUTLER} has recently revolutionized the state of the art, providing models which do not require fine-tuning as they show impressive zero-shot generalizability~\cite{ye2023gaussian3DRECONSTRUCTION, qin2023langsplatRECONSTRUCTIONLANGUAGE}.

In this work, we propose to overcome the limited generalization capabilities of FSS for robot grasping by presenting a novel approach based on zero-shot segmentation through foundation models~\cite{kirillov2023segmentSAM, wang2023cutCUTLER}, followed by a high-performance few-shot classifier~\cite{hu2022pushingPMF}. 
In our vision, few-shot semantic segmentation can be seen as an emerging behavior of 
%is therefore obtained by means of 
the innovative combination of these two techniques, as the original zero-shot segmentation and FSC models are not designed for this task. Figure~\ref{fig:approach} outlines a high-level representation of the approach: given an RGB-D image and one or more segmented examples of the target object, a zero-shot segmentator is in charge of segmenting all the objects in the input RGB image while the FSC selects the candidate segmentation mask among all the possible masks extracted. To enable the FSC to process the masks of the zero-shot segmentator, we propose to swap the inputs, namely query and support, considering the output of the zero-shot segmentator as the support and the original support as the image to classify. In such configuration, instead of classifying the query into one of the classes represented in the support, we use the latter to select the best query among all the segmentation masks produced.

%%%%%%%%%%%%%%%%% RISCRITTO
%We also demonstrate that the proposed FSS approach can be exploited to build a grasp synthesis model for FSG by extracting the logit heatmaps from the segmentator instead of the masks without requiring any fine-tuning, differently from previous art~\cite{liu2022unseenFSSRG, holomjova2023gsmrGSMR, barcellona2023fsgFSGNET}.
%%%%%%%%%%%%%%% NUOVA VERSIONE
We also demonstrate that the proposed FSS approach can be exploited to obtain a grasp synthesis model for FSG by using the output of the segmentation model as an heatmap encoding the probability of the correct object, similarly to~\cite{barcellona2023fsgFSGNET}.  
%%%%%%%%%%%%%%%%%%%%%%%%
The proposed method was thoroughly validated on two standard benchmarks for robot grasping: Graspnet-1B~\cite{fang2020graspnetGRASPNET} and OCID-grasp~\cite{ainetter2021endOCIDGRASP}, demonstrating that the few-shot segmentation of our method overcomes the State Of The Art (SOTA) under such standard experimental conditions. 
%%%%%%%%%%%%%%%%%% RISCRITTO
%The remarkable results in FSS can be used to overcome the generalization limitation highlighted in prior research about few-shot semantic grasping~\cite{barcellona2023fsgFSGNET}.
%%%%%%%%%%%%%%%%%% NUOVA VERSIONE
The improvement in FSS is due to the high generalization of the zero-shot and few-shot models achieved through their ViT backbone trained on a vast dataset, providing a method that overcomes the limitation in the target object identification of prior few-shot semantic grasping approaches~\cite{liu2022unseenFSSRG, holomjova2023gsmrGSMR, barcellona2023fsgFSGNET}.
%%%%%%%%%%%%%%%%%%

%By embedding our FSS method into a grasp synthesis network, we notably improved the performance on FSG-related metrics, particularly in enhancing the selection of the correct target object.

When dealing with robotic applications, the gap between the system performance evaluated on a dataset and those measured in a real experiment is a crucial element to assess the real impact of a new technology. This was investigated by means of a real-world validation of our grasping pipeline using a diverse set of 47 objects. Such objects were chosen to align with the demands of potential application scenarios for few-shot semantic grasping, encompassing items relevant to service robots (\eg spoons, joysticks), warehouse robots (\eg boxes, bottles), and smart manufacturing tools (\eg screwdrivers, hammers). 
%%%%% FRASE SOTTO
%The real experiment also tested the robustness of the approach in choosing the correct target among different instances or when the support has a different visual appearance.
%%%%% RISCRITTO
The experiment in the real scenario also tested the robustness of the approach in choosing the correct instance when there are more objects of the same class or when the support has a different visual appearance compared to the target object in the scene.
%%%%%%%%%%%%%

In summary, the main contributions of this work can be listed as follows: (1) we propose a novel approach to achieve few-shot semantic segmentation, combining an inverted FSC model with a zero-shot segmentator, achieving effective results without fine-tuning the architecture; (2) we employ the proposed FSS module in the context of robotic grasping by embedding its semantic heatmaps into the grasp synthesis; (3) we thoroughly evaluate the performance of the proposed model both for the segmentation and grasping tasks, establishing new state-of-the-art performance; (4) we conducted real-world experiments to demonstrate the robustness of our approach in recognizing new objects, even when the support has a different visual appearance from the target or when different objects with the same semantic are found.

The remainder of the paper is organized as follows. Section~\ref{sec:rw} reviews the works related to few-shot semantic grasping, zero-shot segmentation and few-shot classification. Section~\ref{sec:definitions} gives a formal definition of the few-shot semantic grasping. In Section~\ref{sec:method} the main elements of the proposed approach are described in details. In Section~\ref{sec:experiments}, we describe the evaluation procedure, the dataset used and the real-world experiments. In Section~\ref{sec:results} the results are reported and discussed. Finally, in Section~\ref{sec:conclusions}, conclusions are drawn and future directions of research identified.

\section{RELATED WORKS}
\label{sec:related_works}

\label{sec:rw}
This work takes advantage from the state-of-the-art in multiple fields, namely: few-shot semantic robot grasping, foundation models for zero-shot segmentation and few-shot semantic classification, that are discussed in details in the next paragraphs.

\subsection{Few-shot semantic segmentation for grasping}

% robot grasping
When a robot needs to grasp an object, the position and orientation of the gripper with respect to the object should be determined. This problem is solved by robotic grasping, whose goal is to generate suitable grasping poses for successfully picking objects using a robotic arm equipped with a gripper. Such poses strongly depend on the gripper structure and motion capabilities; while various gripper types are available, two-finger parallel grippers remain the most popular for academic research, and are therefore considered in this study. 
%%%%%%%%%%%%% RISCRITTO
%This popular type of gripper can be represented using either 7 degrees of freedom (7-DoF)~\cite{fang2020graspnetGRASPNET} or 4 degrees of freedom (4-DoF) poses~\cite{ainetter2021endOCIDGRASP}:
%%%%%%%%%%%%%%%%% NUOVA VERSIONE
When using this popular gripper, grasping positions can be represented using either 7 degrees of freedom (7-DoF)~\cite{fang2020graspnetGRASPNET} or 4 degrees of freedom (4-DoF)~\cite{ainetter2021endOCIDGRASP}:
%%%%%%%%%%%%%%%%
the second model, although less versatile, offers a simpler yet effective representation since it can be equivalently modelled as a bounding box~\cite{ggcnn}, and is considered in this work for grasping pose synthesis.

Numerous studies in robotic grasping focus on geometry-based approaches, which lead to remarkable results~\cite{ggcnn, antipodal}, but do not provide information about how to determine which object to grasp. Some works acquire this information thanks to embedded segmentation branches or networks in the grasp synthesis models~\cite{duan2023semanticGRASPSEMANTIC, ainetter2021endOCIDGRASP}. However, as discussed in the previous section, semantic segmentation methods cannot easily generalize to novel objects, therefore few-shot semantic segmentation was introduced in more recent robotic grasping systems: Grasping Siamese Mask R-CNN (GSMR-CNN)~\cite{holomjova2023gsmrGSMR} adapts a one-shot mask R-CNN model by adding a branch to compute the grasping pose as a bounding box; Few-shot Semantic Grasping Network (FSG-Net)~\cite{barcellona2023fsgFSGNET} exploits the FSS named ASGNet~\cite{li2021adaptiveASGNET} and a modular architecture to synthesize a grasp pose for a target object. In~\cite{liu2022unseenFSSRG}, a few-shot semantic segmentation network is used to drive the prediction of the 7-DoF grasps. 
%%%%%%%%%%%%%%%%%%%%%% INIZIO PARTE DA EVENTUALMENTE RIMUOVERE PER FUSIONE
%Despite the promising results, the accuracy of these methods in recognizing the target object with few-shots segmentation is still limited: in this paper, we improve accuracy by exploiting a zero-shot segmentator with a few-shot classifier to enhance the accuracy in predicting the correct target in FSS for robot grasping. 

%\subsection{Few-shot semantic segmentation}
%The task of semantic segmentation aims to predict pixels-level annotation on images, assigning to each pixel a label from a predefined set of classes. Despite the notable results achieved by state-of-the-art solutions~\cite{xie2021segformerSEGFORMER}, a significant limitation of semantic segmentation models is the lack of generalization to new classes without re-training on a new and extensive dataset. To address this issue, few-shot semantic segmentation~\cite{dong2018fewPROTO, peng2023hierarchicalHDMNET} endeavors to generalize to new classes observing only a few labeled examples, known as support set.
%%%%% AGGIUNTO PER COMMENTO SU MANCANZA DI DEFINIZIONE
%Introduced in~\cite{shaban2017oneFSS2017}, few-shot semantic segmentation aims to segment a query image to identify pixels of a class that was not observed at training time, using only one or five labelled examples (shots).
%%%%%%%%%%%%
%%%%%%%%%%%%%%%%%%%%%%%%%% FINE PARTE DA RIMUOVERE PER FUSIONE
%%%%%%%%%%%%%%%%%%% PARTE DA AGGIUNGERE
The main limitation of these approaches is the scarce accuracy in selecting the correct object, caused by the few-shot segmentation models they embody in their methods. 

Introduced in~\cite{shaban2017oneFSS2017}, few-shot semantic segmentation aims to segment a query image to identify pixels of a class not used during training, using only one or five labelled examples (shots). 
%%%%%%%%%%%%%%%%%%%
The prevalent approaches to solve the FSS problem are metric-based~\cite{tian2020priorPFENET, zhang2021selfCANET}, which means that a feature set extracted by an encoder, called embedding, is mapped into a metric space where the distance serves as a measure of similarity between the support set and the query image.
Among the metric-based approaches, prototype-based methods are the most commonly used~\cite{dong2018fewPROTO, wang2019panetPANET, li2021adaptiveASGNET}: they learn the prototype of a new class using the embeddings of the shots in the support set in the metric space. The prototype is subsequently matched with the embeddings of the query image to predict the segmentation mask~\cite{dong2018fewPROTO, wang2019panetPANET}. Some approaches proposed associating multiple prototypes for each class~\cite{li2021adaptiveASGNET} or accumulating knowledge from previously seen classes~\cite{sun2022singularBAM, peng2023hierarchicalHDMNET}. 
In contrast to the models discussed above, we introduce a novel approach for few-shot semantic segmentation that exploits a zero-shot segmentator combined with a few-shot classifier.
%%%%%%%%%%% DA AGGIUNGERE IN CASO DI FUSIONE
This method, when employed for the few-shot semantic grasping, is able to improve the selection of the target object thanks to its generalization capabilities.  
%%%%%%%%%%%%%%%%%%%%%
%Neither of these components is designed explicitly for few-shot segmentation. However, they achieve impressive results in this task when fused together.

\subsection{Foundation models for zero-shot segmentation}
Foundation models typically are promptable models trained on large-scale data, showcasing impressive zero-shot generalization capabilities and making them suitable as a base for novel tasks without additional training or fine-tuning~\cite{zhao2023surveyLLMSURVEY}.
Recently, models specifically designed for computer vision tasks have been proposed, such as the Segment Anything Model (SAM)~\cite{kirillov2023segmentSAM}. This architecture consists of an image encoder and a prompt encoder, which, respectively, map pixels and the user's prompt on the image into the same feature space: leveraging this shared representation, the model's decoder reconstructs a segmentation mask that aligns coherently with the provided prompt. SAM's training strategy follows a ``model-in-the-loop'' approach, starting with manual annotations assisted by the model and progressing towards a self-supervised and fully automatic phase, enabling scaling of the dataset size.
SAM has demonstrated strong generalization capabilities by taking advantage of user-provided prompts, such as points or bounding boxes, along with the vast knowledge accumulated during the training phase. SAM is extremely general-purpose and demonstrated to be easily adaptable to a variety of different applications: several works already built upon SAM, exploiting its zero-shot segmentation capabilities~\cite{ye2023gaussian3DRECONSTRUCTION, saviolo2023unifyingSAMTRACKING, nanni2023improvingSAMZS}, positioning it among the most relevant foundational models in computer vision.
Another model that demonstrated promising zero-shot generalization capabilities is Cut-and-LEaRn (CutLER)~\cite{wang2023cutCUTLER}. CutLER is trained on the ImageNet dataset in an unsupervised manner: pseudo masks are extracted using a Vision Transformer model (ViT)~\cite{dosovitskiy2020imageVIT} pre-trained through DINO~\cite{caron2021emergingDINO} without any human supervision. ViT's self-attention mechanism~\cite{vaswani2017attentionTRANSFORMER} estimates rough segmentation masks using the correlation between patches in the image, subsequently exploited as pseudo-labels. This approach allows the generation of a massive dataset to train the segmentator in a self-supervised fashion at the cost of a rather loss of quality of the pseudo-labels, which is balanced thanks to a custom loss function used for training.
In the proposed approach, we leverage a foundation model with zero-shot capabilities to extract high-quality segmentation masks from the query image, achieving new state-of-the-art performance without fine-tuning the model on specific datasets.

\subsection{Few-shot classification}
Similarly to FSS, few-shot classification aims to classify a query image among N unseen classes (ways) that are not represented in the training set, but rather defined by means of K examples (shots) per class~\cite{snell2017prototypicalPROTONET}.
One difference between FSS and FSC is that FSC targets N-ways, whereas FSS usually focuses on 1-way \cite{okazawa2022interclass1way1, boudiaf2021fewNOMETA}, 
%%%%%% RISCRITTO
%meaning only one class per inference can be given in the support. 
%%%%%% NUOVA VERSIONE
meaning that the support set represents only one new class to segment.
%%%%%%
As with FSS, the most prominent works are based on metric~\cite{sung2018learningFSCMETRIC, wertheimer2021fewRECONSTRUCTION, hiller2022rethinkingFEWTURE} and prototype learning~\cite{snell2017prototypicalPROTONET, hu2022pushingPMF}. Recently, Hu~\etal~\cite{hu2022pushingPMF} demonstrated that adopting the Pretraining-Metalearning-Finetuning (PMF) paradigm can significantly increase the accuracy of FSC. The authors pre-trained a ViT model~\cite{dosovitskiy2020imageVIT} with DINO~\cite{caron2021emergingDINO} on massive datasets and, via meta-learning, transferred the knowledge to a simple prototype learner, that was then fine-tuned in case of domain gaps. In this work, we propose to exploit the remarkable performance of PMF to score the segmentation masks from a zero-shot segmentator. %by inverting the role of the support and the query.

\section{DEFINITIONS}
\label{sec:definitions}
\label{sec:definitions}
\begin{figure*}[!t]
\centering
\resizebox{0.99\linewidth}{!}{
%\framebox{\parbox{3in}{Add image showing skeletons VS 3D Segmentation for qualitative comparison}}
\includegraphics[]{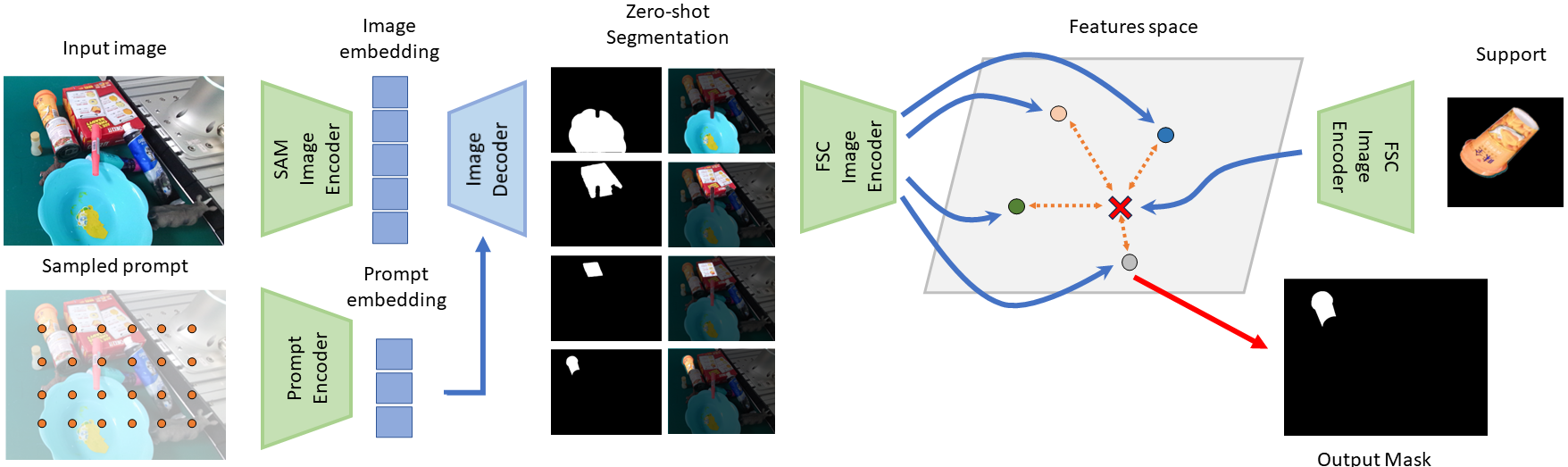}
}
\caption{The few-shot semantic segmentation approach introduced in this work. The input image is given in input to a zero-shot segmentator (\eg SAM). During the decoding, the features extracted from the grid sampled prompts are injected. The produced masks are used to filter the possible candidates. Each candidate is mapped into the feature space of a few-shot classifier together with the support features. The most similar features are used to extract the segmentation mask.}
%Colors distinguish different layer type, while numbers indicates the number of output channels from convolutional layers.
%(i.e. double Conv2D, Conv2D, MaxPool, Upsampling)
\label{fig:fewshot_network}
\end{figure*}

%Before proceeding with a detailed description of the grasping system, some definitions related to few-shot semantic segmentation and grasping need to be given.

Before proceeding with a detailed description of the methodology, formal definitions of few-shot semantic segmentation and grasping are given.

% definizione di few-shot semantic segmentation = un support S è un inseme di tuple (I, M) che data una immagine I_q
\textbf{Few-shot Semantic Segmentation.}  We define a support set as $\mathcal{S} = \{ (I_s, M_s)_i \, , \, i = 1,\cdots,k \} $, that is a set of $k$ tuples of images $I_s$ and related segmentation masks $M_s$ highlighting a class.
The goal of FSS is to estimate the segmentation mask $M_q$ of a given query image $I_q$ for the class described by $\mathcal{S}$ that was not seen during training.
%In this setting, we define the training set as $Sp_{train} = \{ (I_q, M_{q}, S)_i \, , \, i=1,...,n \}$ and the test set as $Sp_{test} = \{ (I_q, M_{q}, S)_i \, , \, i=1,...,m \}$. The two sets have no class in common, formally $\mathrm{cls}(Sp_{train}) \, \cap \, \mathrm{cls}(Sp_{test}) = \{ \emptyset \} $, where  $\mathrm{cls}( \cdot )$ return the list of classes in a set.
%%%%%%% AGGIUNTO
%The episodic learning described above has repercussions on the training and testing disposition.
To solve FSS, the learning phase is often configured as an episodic learning. 
%%%%%%%
We define the training set as $\mathcal{D}_{train} = \{ (I_q, M_{q}, S)_i \, , \, i=1,\cdots,n_{tr} \}$ and the test set as $\mathcal{D}_{test} = \{ (I_q, M_{q}, S)_i \, , \, i=1,\cdots,n_{te} \}$. The sets must adhere the condition $\mathrm{cls}(\mathcal{D}_{train}) \, \cap \, \mathrm{cls}(\mathcal{D}_{test}) = \{ \emptyset \} $, where  $\mathrm{cls}( \cdot )$ return the list of classes in a set. This means that training and testing have no classes in common.

\textbf{Grasping.}  Following many works~\cite{ggcnn, ainetter2021endOCIDGRASP}, and considering a two finger parallel gripper, the grasp described in the 4-DoF representation takes the form:
%described using 4-DoF poses, we define a grasp as the tuple
\begin{equation}
\label{eq:grasp}
    gr = (x, y, \theta, w)\,,
\end{equation}
where $x$ and $y$ are the coordinates of the gripper, $\theta$ is its orientation along the axis perpendicular to the image plane, and $w$ is the width of the gripper. It follows that the grasp can be represented as an oriented bounding box.  %\textcolor{blue}{Figure []} shows an example of grasp on an object.

\textbf{Few-shot Semantic Grasping.} Following~\cite{barcellona2023fsgFSGNET}, given a query-support tuple $ (I_{q}, \mathcal{S})$, we define the Few-Shot Semantic Grasping as the task of inferring the optimal grasp defined as:
\begin{equation}
    gr_{s}^* = (x^*, y^*, \theta^*, w^*)\, \;\; s.t. \;\; \mathrm{cls}(\{I_q(x^*,y^*)\}) = \mathrm{cls}(\mathcal{S})\,,
\end{equation}
where $ I_q \in \mathbb{R} ^{4{\times}H{\times}W}$ is the RGB-D query image and $\mathcal{S} = \{ (I_s, M_s)_i \, , \, i = 1,\cdots,k \} $ 
%%%%%% Riscritto
%comprises the RGB support images  $I_{s} \in \mathbb{R} ^{k{\times}3{\times}H{\times}W}$ and the corresponding binary segmentation masks for the target class $M_{s} \in \mathbb{R} ^{k{\times}H{\times}W}$.
%%%%%%% Nuova versione
comprises the support images as previously defined.

\section{METHOD}
\label{sec:method}
\label{sec:method}
To address the Few-Show Semantic Grasping problem introduced in Section~\ref{sec:introduction}, we propose to merge the zero-shot capabilities of foundation segmentation models along with the abilities of few-shot image classifiers to achieve a high-quality semantic mask for unseen classes. As described in Sec.~\ref{sec:definitions}, given a query image $I_q$ and a support set showing examples of a class that was not part of the training set, the objective of FSS is to compute the best estimation of $M_q$, the segmentation mask of $I_q$. 
%%%% RISCRITTO
%Figure~\ref{fig:fewshot_network} outlines the proposed approach: initially, the zero-shot segmentator extracts a set $(M_q^0,\cdots,M_q^m)$ of possible masks from $I_q$.
%=== occhio qui, ho riscritto senza cancellare l'originale
%Figure~\ref{fig:fewshot_network} outlines the proposed approach: in the first stage, the zero-shot segmentator composed of the SAM Image Encoder followed by a decoder extracts a set $(M_q^0,\cdots,M_q^m)$ of possible masks from the input image.
%=== fine
%In the second stage, the masks are fed to the few-shot classifier to score them against the support set. Finally, the segmentation mask with the highest similarity score to the support is chosen as $M_q$. 
%%%%%%%  NUOVA VERSIONE
Figure~\ref{fig:fewshot_network} outlines the proposed approach using the Segment Anything Model (SAM) as zero-shot segmentator. In the first stage, the query image and a sampled prompt are processed by two different encoders, whose outputs pass through SAM's decoder module to extract a set $(M_q^0,\cdots, M_q^m)$ of possible masks from the input image. In the second stage, a set of masked images $(Im_q^0,\cdots, Im_q^m)$, produced by filtering $I_q$ with $M_q^i$, are fed to the few-shot classifier to score them against the support set. Finally, the segmentation mask with the highest similarity score to the support is chosen as $M_q$.
Despite the previous example assuming SAM as mask extractor, any zero-shot segmentator can be employed to infer the set of masks  $(M_q^0,\cdots,M_q^m)$ of the query image $I_q$ thanks to the modular architecture of the proposed method. In this work, we considered both SAM and CutLER as mask extractors.
%%%%%
The proposed segmentation architecture can be easily embedded into a few-shot semantic grasping model together with a grasp synthesis module. In this work, we adopted an approach similar to~\cite{barcellona2023fsgFSGNET, ggcnn}, 
%%%%%%% RISCRITTO
%but feeding the semantic information and removing the backbone.
%%%%%% NUOVA VERSIONE
but directly feeding the semantic, encoded into the heatmap $H_q$ obtained before discretizing the output of the segmentator into a mask, as part of the input to the network while removing the backbone module.
%%%%%

The following subsections delve into the details of the unsupervised segmentator, the few-shot classifier, and the grasp synthesis network.

\subsection{Zero-shot segmentator}
The first step involves providing a reliable segmentation of the query image using a zero-shot approach. In this work, we use two segmentators: SAM and CutLER. SAM is composed of three blocks: an image-encoder $G$, a prompt-encoder $P$ and a mask-decoder $D$. $G$ is a ViT-H/16 with 14$\times$14 windowed attention and four equally-spaced global attention blocks trained in a self-supervised method via masked autoencoders \cite{he2022maskedMASKED}; it maps the input image into an embedding of dimension $C \times H \times W$ with $C=256$, $H=64$ and $W=64$.
$P$ uses the point prompt to produce a 256-dimensional vector containing the positional encoding of the location~\cite{tancik2020fourierPOSITIONALENCODING} and two learnable parameters to distinguish foreground and background points. 

To use SAM as a zero-shot unsupervised segmentator, a grid of single-point inputs sampled over the image is used as input to $P$ to produce the set of masks $(M_q^0,\cdots,M_q^m)$. 
The masks are produced by the decoder $D$, which modifies the standard Transformer decoder~\cite{vaswani2017attentionTRANSFORMER}, to produce masks by exploiting self-attention to the prompt embeddings and cross-attention between the prompt embeddings and the image embeddings. Finally, masks are filtered to enhance quality and to remove duplicates using non-maximal suppression. 

CutLER uses a Mask-RCNN model trained using a pipeline proposed by the authors. In particular, the self-supervised masks were extracted by ViT pre-trained with DINO~\cite{caron2021emergingDINO} and MaskCut~\cite{wang2023cutCUTLER}. Finally, the model underwent a self-training stage to refine the prediction.
The segmentation masks $M_q^i$ produced by SAM and CutLER are applied to $I_q$, turning detected objects into a set of masked images $(Im_q^0,\cdots,Im_q^m)$.

\subsection{Few-shot classifier}
In our approach, we use the few-shot image classifier to determine the masked object $Im_q^i$ produced by the zero-shot segmentator most semantically similar to the support set.
By definition, a FSC model associates a query image to one of the N possible classes, namely the N-ways. In our context, the support contains examples of the target class only, configuring a 1-way problem, that is ill-posed for a FSC. 
%Thanks to the segmentation masks of the support, it is possible to separate the object from the background, designing the classification as a 2-way task and then classifying each masked image produced by the zero-shot segmentator. This configuration would lead to the creation of a prototype for the background, which includes all the features that are not related to the desired class. However, generating a representative prototype from only masked images is not achievable due to the high variability of this class, which is not well-represented by one or five shots. 
Designing the problem as a 2-way task is achievable by creating a prototype of the background, but this choice would bring low performance due to the high variability of such a prototype,  which may not be adequately expressed by the few support images.  

The proposed solution is inverting the role of the support set and the masked images (our query images) to avoid modelling the background. The support is mapped into a prototype in the same embedding space of the masked images and then classified to the most similar embedding. This approach allows retrieving the predicted mask $M_q$. When the method is used in the grasp synthesis, the heatmap $H_q$, obtained before discretizing the output of the segmentator, is extracted instead of $M_q$. 

In our experiments, we used PMF~\cite{hu2022pushingPMF}, a state-of-the-art model for few-shot image classifier that adapts a ViT~\cite{dosovitskiy2020imageVIT} model to the task with a prototypical approach. The model extracts the features from the support examples and creates a vector, named prototype, for each class. The query features are mapped into the same space and classified by applying a similarity score (cosine similarity) to the prototypes.

%%%%%%%%%%%%%%%%%%%%%%%%%%%%%%%%%%%%%%%
\begin{figure*}[ht!]
\centering
\resizebox{0.99\linewidth}{!}{
%\framebox{\parbox{3in}{Add image showing skeletons VS 3D Segmentation for qualitative comparison}}
\includegraphics[]{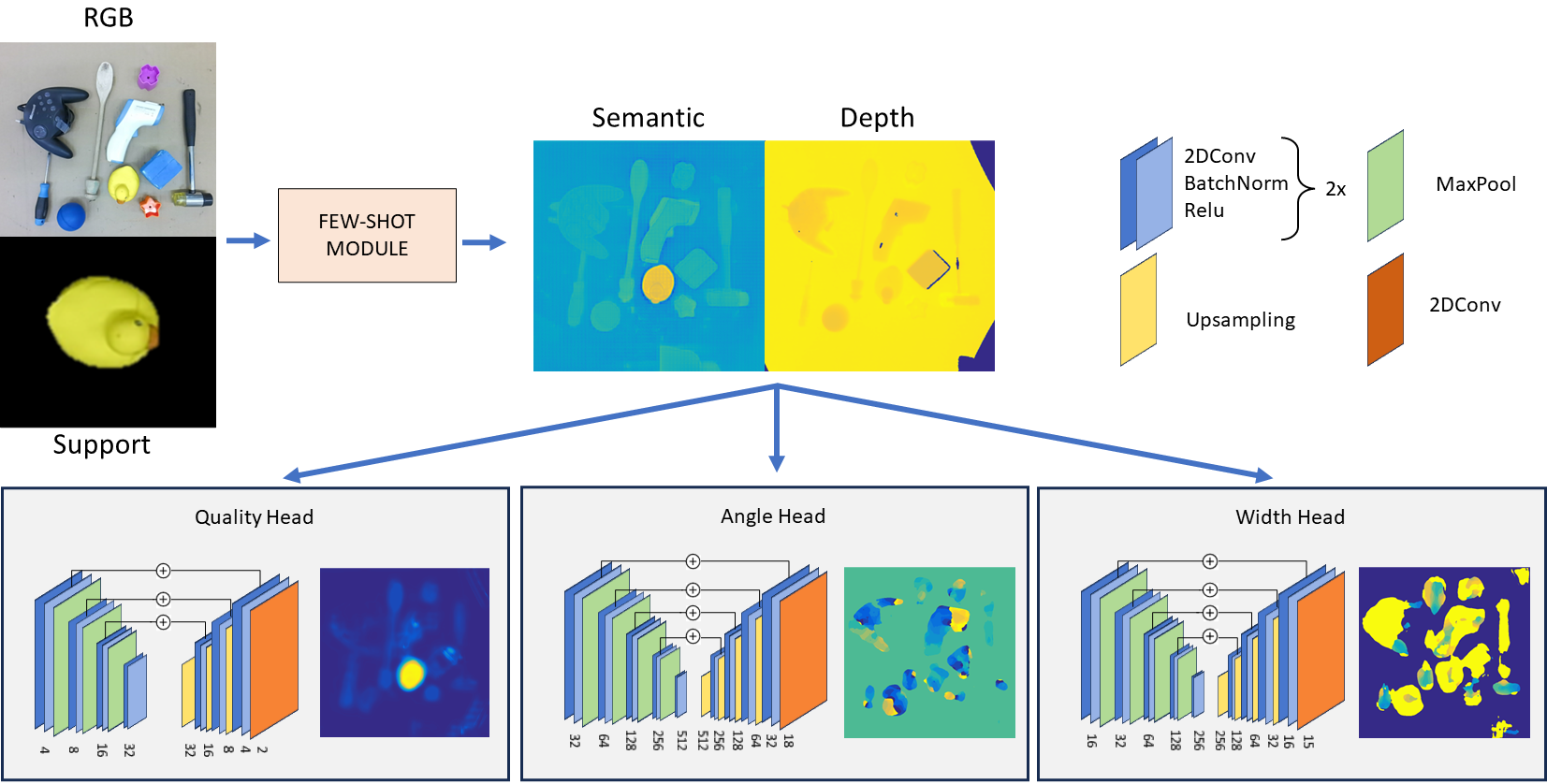}
}
\caption{The grasp synthesis network. The semantic heatmap is extracted starting from the RGB and support image. Then, it is concatenated with the depth image to compose the input for three modules: Quality Head, Angle Head and Width Head. These modules produce three heatmaps encoding the grasp angle, width and position.}
%Colors distinguish different layer type, while numbers indicates the number of output channels from convolutional layers.
%(i.e. double Conv2D, Conv2D, MaxPool, Upsampling)
\label{fig:grasping_network}
\end{figure*}

\subsection{Grasp synthesis network}
\label{sec:grasp_synth}

The grasp synthesis network is depicted in Figure~\ref{fig:grasping_network}. The grasp, formally defined in (\ref{eq:grasp}), is encoded into three heatmaps: $Q, A$ and $W$, predicted by three UNet~\cite{unet} heads, representing the quality, the angle and the width of the predicted grasp ($gr_p$) as in~\cite{ggcnn}. The width and angle modules predict the aperture and the orientation of the gripper, whereas the quality head is used to infer the center point of the grasp. Information about the target object is injected through
%%%% AGGIUNTO
the heatmap $H_q$ produced by
%%%%
the proposed few-shot segmentator, synthesized through the cascade of a zero-shot segmentator followed by a few-shot classifier. The output of this module is concatenated with the depth image to mix semantic knowledge with geometric information, similarly to~\cite{barcellona2023fsgFSGNET}, and provided as input to the three modules. 
%%%% RISCRITTO
%The quality head proposed incorporates an increased channel dimension (out channels $[4,8,16,32]$) compared to~\cite{barcellona2023fsgFSGNET} because the segmentation is more reliable. The backbone is also missing because semantics can accurately distinguish the target object, and keeping the former would increase the computational burden.
%%%%%% NUOVA VERSIONE
The proposed model is inspired by~\cite{barcellona2023fsgFSGNET} with two main differences: the quality module has an increased channel dimension (decoder channels $[4,8,16,32]$), and the backbone is removed. 
Both these changes are due to the increased reliability of the segmentation, which offers a valid localization of the target object. 
%%%%%%%%%%%

The grasp pose is inferred from the heatmaps. The quality 
%%%%%%%%%%
%module outputs two channels
%%%%%%%
heatmap has two channels,
%%%%%%
one for the foreground, \ie suitable grasping location for the target object, and one for the background. To compute the final grasp prediction $gr$, as described in (\ref{eq:grasp}), we applied a softmax operator to the foreground heatmap $Q_f$, bounding the output of each pixel in the range $[0, 1]$. 
%%%%%%%%%% RISCRITTO
%We then extracted the maximum peak with the highest value in the quality heatmap together with the surrounding pixels with a value higher or equal to 95\% of the maximum. We kept the center $(x,y)$ of such peak. When two or more picks share the same dimension we sampled one among them.  
%Finally, we extract the values of the angle prediction $\theta = A(x,y)$ and the width prediction $w = W(x,y)$ at the position $(x,y)$, where $A$ is the generated angle heatmap and $W$ the width heatmap.
%%%%%%%%%%% NUOVA VERSIONE
%We then extracted the maximum peak with the highest value in the quality heatmap together with the surrounding pixels with a value higher or equal to 95\% of the maximum. 
We then extracted the peaks with the highest value in the quality heatmap with their surrounding pixels having a value higher or equal to 95\% of the maximum value in the heatmap. We selected the peak with the broadest extension as the final prediction.  
When two or more picks share the same dimension we randomly sampled one among them. The center $(x,y)$ of the peak corresponds to the center of the grasp, and is also used to extract the angle prediction as $\theta = A(x,y)$ and the width prediction $w = W(x,y)$, where $A$ is the generated angle heatmap and $W$ the width heatmap.

\section{EXPERIMENTS}
\label{sec:experiments}
\label{sec:experiments}
%%%%%%%%%%%%%%%%%%%
% MATTEO
% Qui inserirei un paragrafo con l'overview degli esperimenti e considerazioni sul training, stressando cosa gli esperimenti vogliono dimostrare (generalizability of the approach)

% 1) Considerazioni sul training. Dire che intero vostro sistema punta ad essere approccio generale al robot grasping, ottenuto in due modi: zero-shot (SAM) che non richiede ulteriore training o fine-tuning, grasp synthesis con i suoi 3 moduli allenati su GraspNet come generico dataset di grasp

% 2) Esperimenti mirato a dimostrare generalizzazione approccio a vari scenari senza uso di fine-tuning, in particolare considerando valutazione su dataset letteratura (GraspNet e Ocid) e validazione sperimentale con robot reale 

%%%%%%%%%%%%%%%%%%%
In this section, provide details about the training process and we describe the experiments conducted to validate the proposed approach. As previously emphasized, the few-shot segmentation module does not require any fine-tuning: even though it could potentially enhance performance, it would also specialize the models to the training setup, leading to a loss of generality. On the contrary, our goal is to demonstrate the generalizability of the approach to the context of robot grasping without specializing the models. 
%%%%%%%%%%%% RISCRITTO
%Conversely, the modules related to grasping are trained on Graspnet-1B and Ocid-grasp, similarly to~\cite{barcellona2023fsgFSGNET}.
%%%%%%%%%%%%% NUOVA 
While the modules for segmentation do not need training on grasping datasets, the modules related to grasp synthesis are trained on Graspnet-1B and Ocid-grasp, similarly to~\cite{barcellona2023fsgFSGNET}.
%%%%%%%%%%
The experiments primarily focus on the Graspnet-1B~\cite{fang2020graspnetGRASPNET} and Ocid-grasp~\cite{ainetter2021endOCIDGRASP} datasets 
%to provide a comprehensive evaluation of the segmentation performance of the proposed few-shot semantic segmentation module and the grasp accuracy of the grasp synthesis network. These datasets 
contain characteristics, such as multiple objects in the scenes and annotations for both grasp and semantics, which are essential for evaluating the proposed approach.
%%%%%%%% AGGIUNTO PER COMMENTO
Therefore, the two datasets were used to provide a comprehensive evaluation of both the proposed few-shot semantic segmentation module and the grasp synthesis network. 
%%%%%%%
Additionally, experiments in the real world are presented. These experiments evaluate the model's ability to handle a set of different objects compared to those used in the two datasets. Furthermore, they measure how well the few-shot semantic grasping model can select a specific instance and recognize objects even when the support contains a different instance.

\subsection{Training}

As explained in Section~\ref{sec:method}, the grasp synthesis network comprises three blocks, which were trained separately following the approach in~\cite{barcellona2023fsgFSGNET}.
Both for Graspnet-1B and Ocid-grasp, the ground truth is provided in the form of planar grasps, \ie defined by a set of bounding-boxes. However, our system works using a heatmap-based representation, therefore a conversion was needed: we generated the ground-truth heatmaps from the annotated planar grasps by starting with blank images, then considering the annotated grasp $g_{gt}=(x_{gt},y_{gt},a_{gt},w_{gt})$ and, for each one, adding a contribution to the heatmap in the form of a small patch centered at $(x_{gt},y_{gt})$ with a value of $a_{gt}$ for the angle heatmap and $w_{gt}$ for the width heatmap. The angle $a_{gt}$, ranging in $[-90, 90)$ degrees, was discretized into 18 bins to evaluate the angle by solving a classification task. For Graspnet-1B, the grasp width $w_{gt}$ falls within the range $[0,150]$ millimeters and was discretized into 15 classes. Ocid-grasp lacks a direct correspondence between the width of the gripper and the width of the bounding box, therefore we considered dimensions in pixels in the range $[0, 90]$ discretized into 15 classes.
To generate the ground truth for the quality module, we applied the same procedure weighting the boxes of the target object with a value of 1 to define the positions that can be grasped. %
% diciamo che abbiamo trainato sul training usando le CE loss
During training, we employed the cross-entropy loss for each module. For the angle and width modules, the loss was weighted as outlined in~\cite{barcellona2023fsgFSGNET}
%%%% AGGIUNTO
to recover from the high presence of background pixels that produce an unbalanced task.
%%%%%%%%%
All the modules were trained on an NVIDIA Titan RTX with 24 GB of VRAM. Due to GPU memory constraints, SAM predictions on the training data were extracted offline before training.

\subsection{Datasets and metrics}
% dataset utilizzati
%To validate the approach for segmentation and grasping we utilized Graspnet1B~\cite{fang2020graspnetGRASPNET} and Ocid-grasp~\cite{ainetter2021endOCIDGRASP} datasets. %We also tested the approach in a real world setting with more than 40 objects never seen during training. 

When applied to the data in Graspbet1B and Ocid-grasp, SAM tends to generate a high number of segmentation masks, which should be reduced to control the complexity of the segmentation data while retaining relevant information for the grasping task and keeping computational time under control. This was done
%When dealing with these datasets, using all the segmentation masks automatically predicted by SAM is challenging because of the high cardinality of the masks. Therefore, we employed a simplification strategy
by selecting the mask 
%%%%%% RISCRITTO
%with the maximal dimension 
%%%%%%% 
with the highest number of pixels
%%%%%%%
as an estimation of the plane where the objects are placed. We discarded all small masks (less than 0.3\% of the image) or those located outside the plane estimated from the largest mask.
%This approach helped to reduce the complexity of the segmentation data while retaining relevant information for the grasping task.

In the following, details of the two datasets previously mentioned
%%%%%AGGIUNTO
and the specific metrics used to evaluate the approach 
%%%%%%%
are reported.
\subsubsection{Graspnet Dataset}
% graspnet com'è composto, cosa abbiamo utilizzato
Graspnet-1B comprises images captured using a Kinect Azure and a RealSense D435 sensors, including 88 different objects arranged in 190 small cluttered groups. These groups are divided into four sets: \textit{train}, \textit{validation}, \textit{similar} and \textit{novel}.
%%%%%%%%%%% AGGIUNTO
The authors provide the division to train and validate grasp synthesis models on the first two sets while testing it with objects similar to the training ones (\textit{similar} set) or completely new ones (\textit{novel} set).
%%%%%%%%%%%
In our experiments, we opted to use the images captured with the Kinect to maintain consistency with the experiments conducted in the real scenario. 
%%%%%%% RISCRITTO
%To evaluate the few-shot semantic segmentation in Graspnet-1B, we employed the mean Intersection over Union (IoU) as in~\cite{liu2022unseenFSSRG}, which considered the train and validation splits.
%%%%%%%%% NUOVA VERSIONE
To evaluate the few-shot semantic segmentation in Graspnet-1B we considered the division proposed in~\cite{liu2022unseenFSSRG}. In this arrangement of Graspnet-1B, namely Graspnet10i, objects from training and validation sets are grouped into four sets of 10 objects. The objective is training on three sets and testing on the remaining. We exploited the same division to test our few-shot segmentation approach, ensuring a fair comparison with other existing methods.
%%%%%%%%%%%%%%
For evaluating the grasps generated by the network, we used the \textit{similar} and the \textit{novel} sets since they contain objects that were never seen during training, aligning with the methodology in~\cite{barcellona2023fsgFSGNET}.
To evaluate if a predicted grasp $gr_p = (x_p, y_p, a_p, w_p)$ is correct, it is compared against the ground truth $gr_{gt} = (x_{gt}, y_{gt}, a_{gt}, w_{gt})$ using two measures, i) Intersection over Union (IoU); ii) difference of the grasp angles; under the following constraints:

%A grasp $g_p = (x_p, y_p, a_p, w_p)$ prediction is considered correct based on two constraints considering its Intersection over Union (IoU) with the ground truth and the difference of the grasp angles:
%is considered correctly predicted if its 
%if the Intersection over Union (IoU) of the bounding-box
%if it satisfies the following constraints on the Intersection over Union (IoU) of the bounding-box and the angle of the grasp when compared to a ground-truth grasp $g_{gt} = (x_{gt}, y_{gt}, a_{gt}, w_{gt})$:
\begin{equation}
    \frac{\mathrm{Intersection}(gr_p, gr_{gt})}{\mathrm{Union}(gr_p, gr_{gt)}} > 0.25\,,
    \label{eq:IoU}
\end{equation}
\begin{equation}
    | a_p - a_{gt} | < 30^{\circ}\,.%°
    \label{eq:a_p_gt}
\end{equation}
This formulation for a correct grasp is widely adopted in literature for bounding box representations~\cite{ainetter2021endOCIDGRASP, holomjova2023gsmrGSMR, ggcnn}. In this study, the height of the bounding box is computed as $w/2$.
Based on this definition, we consider two metrics that take into account the correctness of the target object. The first one is the accuracy in selecting the correct object $A_{sem}$, which checks if the center of the grasp is inside the ground-truth grasp of the target object, represented by the support set. The second one is the accuracy in grasping the target object $A_{semGR}$, which considers the predicted grasp $gr_p$ correct only if the two constraints (\ref{eq:IoU}) and (\ref{eq:a_p_gt}) are satisfied for a grasp on the target object. Tests were performed for 10,000 iterations.

\begin{figure}
  \centering
  \resizebox{0.99\columnwidth}{!}{
  \includegraphics[]{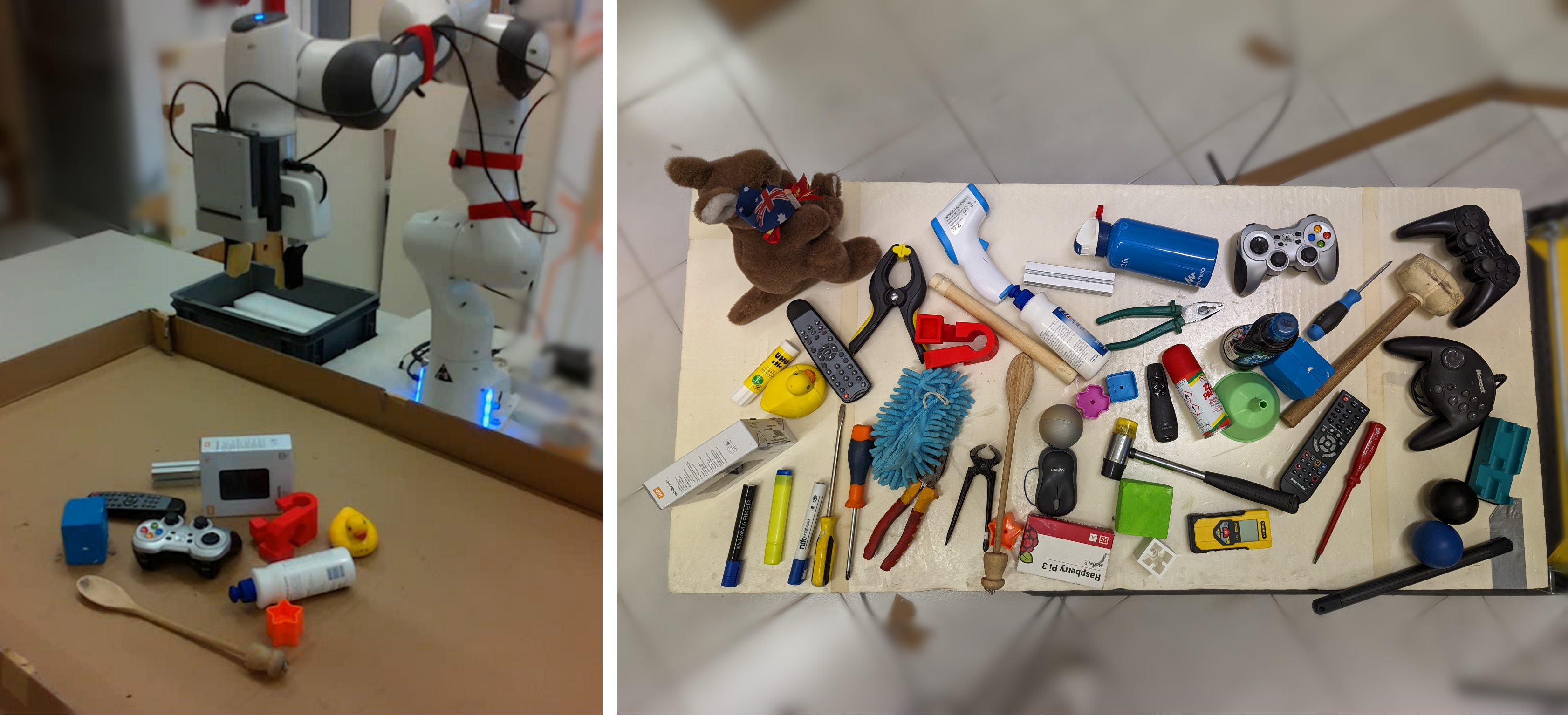}
  }
  \caption{The robotic setup (left) and the set of objects (right).}
  \label{fig:setup}
\end{figure}

\subsubsection{Ocid Dataset}
% ocid com'è composto, cosa abbiamo utilizzato
The Ocid dataset~\cite{suchi2019easylabelOCID} comprises 96 cluttered scenes captured by two ASUS-PRO Xtion cameras with a total of 89 object classes from the Autonomous Robot Indoor Dataset (ARID)~\cite{loghmani2018recognizingARID}  and YCB Object and Model Set~\cite{calli2017yaleYCB}, including RGB-D images and segmentation labels. Ainetter et al.~\cite{ainetter2021endOCIDGRASP} expanded the dataset to include grasping annotations.
% metriche coco e metriche grasp.
For the segmentation evaluation in this dataset, we followed the evaluation procedure outlined in~\cite{holomjova2023gsmrGSMR}, which proposed a one-shot segmentation network for grasping. The authors introduced two train-test splits: image-wise and object-wise. In the image-wise split, the model is tested on new images, while in the object-wise split, the approach is tested with new classes (objects). To be coherent with the few-shot setting, we compared our approach only on the second split, where the models are tested on 8 new classes. In~\cite{holomjova2023gsmrGSMR}, the authors evaluated the segmentation ability of the network with the standard MSCOCO metrics that computes Average Precision (AP) and Average Recall (AR) scores at different Intersection over Union (IoU) thresholds and maximum amount of detections (1, 10, 100). Since our model can compute only one semantic segmentation mask we only compare the AR with 1 detection. We also computed the mIoU for the 8 classes as for the previous dataset. For the grasp evaluation, we computed the $A_{sem}$ and $A_{semGR}$ as previously described.

\begin{figure}
  \centering
  \resizebox{0.99\columnwidth}{!}{
  \includegraphics[]{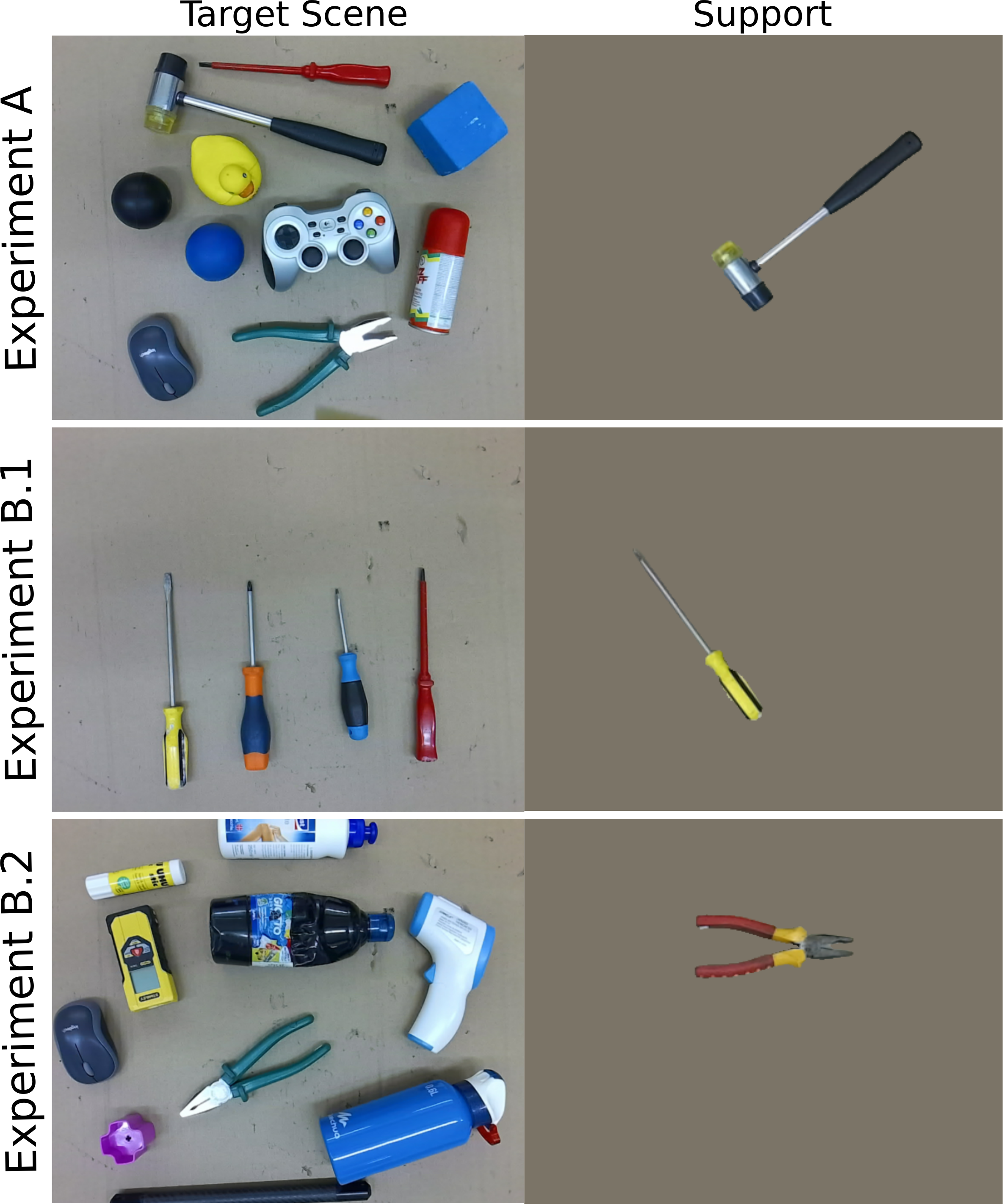}
  }
  \caption{The three real-world experimental setups. Experiment A (top) follows the previous test approach~\cite{barcellona2023fsgFSGNET}. Experiment B.1 (center) test the specificity \ie the capability of recognizing a specific object inside a group of the same class. Experiment B.2 (bottom) investigates the generalizability \ie the ability to recognize the target given a different object of the same class as support.}
  \label{fig:real_world_exp}
\end{figure}

\subsection{Real-world setup} 
The real-world experiment aimed to test the model's ability to bridge the gap between the dataset and real-world scenarios, using 47 objects not comprised in the datasets 
%%%%%%%%%% AGGIUNTO
under the 1-shot and 5-shot settings.
%%%%%%%%%
All the experiments were performed using a Franka Emika Panda manipulator equipped with a Kinect Azure camera calibrated using~\cite{evangelista2022unifiedCALIBRATION} mounted on the wrist of the robot. Fig.~\ref{fig:setup} shows the setup and the  set of 47 objects used for the experiments:
%different from the ones used in the two datasets.
for each one, a support set of 7 images was collected and annotated. The grasps were inferred using the proposed grasp synthesis model on an NVIDIA Titan RTX with 24 GB of VRAM.

Two experiments were designed. Experiment A follows a logic similar to previous works~\cite{barcellona2023fsgFSGNET}:
%, see Figure~\ref{fig:real_world_exp}. 
we selected 10 random objects to place on the table; among them, one was chosen as the target---one or five support images of that object were then picked and presented to our grasping model, which generated the grasping pose from the RGB-D image of the current scene.
%RGB-D image of the current scene to generate a grasping pose with the grasping model.
The predicted planar grasp $gr_{p}$
%is
was then re-projected into camera 3D coordinates using the depth image, and expressed in the robot base reference frame using the hand-eye calibration. Finally, the robot performed the pick. We repeated the procedure for each object on the table; to make all the runs comparable, the previously picked object was put back on the table. We performed 10 repetitions of experiment A for each object, ending up with 100 real-world grasp trials. 
%

%This evaluation setting left the following questions:
Experiment A did not cover two special cases:
(1) the target object is surrounded by objects of the same category; (2) the target object is represented through a support of the same category but with different visual appearance.
To investigate these situations, we designed experiment B
%%%%%%% RISCRITTO
%set up by selecting 7 groups of objects belonging to the same semantic class
%%%%%%%%%%% NUOVA VERSIONE
by organizing some of the 47 objects into seven semantic groups with elements of the same class 
%%%%%%%%%%
but with different appearances: $3\times$ joysticks, $3\times $ markers,  $3\times $ pliers,  $3\times $ remotes, $4\times $ screwdrivers, $3\times $ rubber balls, $4\times $ plastic cubes. In experiment B.1 (instance-level test), we asked the robot to recognize and pick a specific object inside its semantic group, providing the actual target as support. 
We performed 23 runs, one for each object. In experiment B.2, we asked the robot to recognize and pick a target object among a random clutter, providing  a different object of the same category as support (class-level test). We performed one run for every possible combination of target and support inside a semantic group, for a total of 54 repetitions.

\section{RESULTS}
\label{sec:results}
\label{sec:results}
\begin{table*}[t!]
\caption{Result on the Graspnet-10i~\cite{liu2022unseenFSSRG} for the 1-shot and 5-shot settings. The proposed approach has reached the new state of the art in almost all the splits.}
\label{tab:graspnet10i}
\begin{center}
\resizebox{0.99\linewidth}{!}{
\begin{tabular}{|c|ccccc|ccccc|}
\hline
  & \multicolumn{5}{c|}{\rotatebox[origin=c]{0}{\color{black} 1-shot}} & \multicolumn{5}{c|}{\rotatebox[origin=c]{0}{\color{black} 5-shots}}\\
\cline{2-11}
 \multirow{-2}{*}{ Model} & split-0 & split-1 & split-2 & split-3 & mean & split-0 & split-1 & split-2 & split-3 & mean\\
\hline
PaNet~\cite{wang2019panetPANET}          & 21.3 & 32.7 & 26.3 & 19.5 & 25.0 & 27.3 & 39.6 & 33.2 & 25.1 & 31.3 \\
CANet~\cite{zhang2021selfCANET}           & 22.5 & 34.2 & 33.5 & 24.3 & 28.6 & 29.6 & 41.0 & 37.9 & 31.6 & 35.0 \\
PFENet~\cite{tian2020priorPFENET}          & 25.3 & 43.6 & 34.4 & 31.7 & 33.8 & 31.4 & 48.8 & 41.8 & 36.3 & 39.6 \\
CWT~\cite{lu2021simplerCWT}             & 27.1 & 42.8 & 36.2 & 29.7 & 34.0 & 33.5 & 46.4 & 45.7 & 40.5 & 41.5 \\
ASGNet~\cite{li2021adaptiveASGNET}          & 24.5 & 37.5 & 38.1 & 30.1 & 32.6 & 54.1 & 43.3 & 52.0 & 42.8 & 48.1 \\
FSS grasping~\cite{liu2022unseenFSSRG} & 39.2 & \textbf{53.2} & 40.5 & 33.4 & 41.6 & 45.7 & 56.3 & 47.1 & 43.9 & 49.3 \\ 
\hline
Our (CutLER) & 47.5 & 34.2 & 38.0 & 36.8 & 39.1 & 50.9 & 38.3 & 43.3 & 42.1 & 43.7 \\
Our (SAM)    & \textbf{58.9} & 47.3 & \textbf{47.4} & \textbf{44.2} & \textbf{49.6} & \textbf{65.1} & \textbf{56.5} & \textbf{55.0} & \textbf{53.0} & \textbf{57.4} \\
\hline
\end{tabular}
}
\end{center}
\end{table*}

This section summarizes the performance assessment, measured on the two datasets separately for easing the comparison with other works in the literature.

\subsection{Performance assessment on Graspnet-1B}

The tests performed on Graspnet-1B include both evaluation of the proposed few-shot semantic segmentation in Graspnet10i and of the whole grasp architecture on the few-shot semantic grasping task \ie the ability to grasp a target object given few examples. 
%%%%%%%%%%% Da togliere se va bene modifca precedente
%We first evaluated the few-shot semantic segmentator using Graspnet10i~\cite{liu2022unseenFSSRG}. In this arrangement of Graspnet-1B~\cite{fang2020graspnetGRASPNET}, objects from training and validation are grouped into four sets of 10 groups. 
%The objective is training on three sets and testing on the remaining \ie spared objects. We exploited the same division to test our few-shot segmentation approach, ensuring a fair comparison with other existing methods. 
%%%%%%%%%

Our approach 
%, which combines a zero-shot segmentator with a few-shot classifier, 
evidently outperformed the previous state-of-the-art solutions, as visible in Table~\ref{tab:graspnet10i}. It is important to note that our proposed method underwent no training or fine-tuning on the Graspnet10i dataset, clearly showing high generalizability and exploitability in the context of robot grasping. Moreover, the modularity of our method allows for easy embedding and testing of different models. We tested CutLER and SAM as zero-shot segmentator: even though the latter achieves better performance, clearly beyond other solutions, CutLER still provides robust results comparable with the state of the art. 

% bisogna dire che compariamo solo metodi a bounding box --> già detto prima in realtà
To evaluate the grasp accuracy of the whole grasp prediction, we used the \textit{similar} and \textit{novel} sets of Graspnet-1B.
%%%%%%%%%%% Riscritta
%Given the results in Table~\ref{tab:graspnet10i}, we used SAM as a zero-shot segmentator to perform grasp synthesis tests. 
%%%%%%%%%%% Nuova versione
Given the better performance of SAM in every split (Table~\ref{tab:graspnet10i}), in the grasp synthesis we adopted such model as zero-shot segmentator.
%%%%%%%%%%
The resulting metrics are shown in Table \ref{tab:graspnet_grasp}, which compares FSG-Net~\cite{barcellona2023fsgFSGNET}, GGCNN~\cite{ggcnn} and GR-ConvNet~\cite{antipodal} using ASGNet~\cite{li2021adaptiveASGNET} as few-shot module, with our grasp model. The advantages of the novel few-shot segmentation module are clear. The proposed method is capable of increasing $A_{sem}$ \ie the accuracy in choosing the target object, by 13.7\% on the \textit{similar} set and by 9.6\% on the \textit{novel} set. The combination of the novel few-shot module with the proposed architectural changes also positively affects $A_{SemGR}$ \ie the semantic grasp synthesis accuracy, with a gain of 15.82\% and 16.5\% on \textit{similar} and \textit{novel} set, respectively. 
Overall, the test performed on Graspnet-1B confirms that the proposed methods are effective in improving both the few-shot semantic segmentation performance and, consequently, the ability to correctly generate a suitable grasp for a target object compared to previous approaches.

\begin{table}[t]
\caption{Grasp results on the \textit{similar} and \textit{novel} sets of Graspnet-1B~\cite{fang2020graspnetGRASPNET}. $A_{Sem}$ is the accuracy in selecting the correct target object, depicted by the shots. $A_{SemGR}$ considers also the grasp accuracy.}
\label{tab:graspnet_grasp}
\begin{center}

\resizebox{\columnwidth}{!}{
\begin{tabular}{|c|cc|cc|}
\hline
  & \multicolumn{2}{c|}{\rotatebox[origin=c]{0}{\color{black} Similar}} & \multicolumn{2}{c|}{\rotatebox[origin=c]{0}{\color{black} Novel}}\\
\cline{2-5}
 \multirow{-2}{*}{ Model} & $A_{Sem}$ & $A_{SemGR}$ & $A_{Sem}$ & $A_{SemGR}$\\
\hline
GGCNN \cite{ggcnn} + FSS & 41.8 & 30.24 & 43.3 & 34.0 \\
GR-ConvNet \cite{antipodal} + FSS & 34.6 & 23.6 & 40.1 & 28.0\\
FSG-Net~\cite{barcellona2023fsgFSGNET} & 51.5 & 43.5 & 49.4 & 41.0\\
\hline
Our (SAM) 1 shot  & \textbf{59.7}& \textbf{56.8} & \textbf{61.5} & \textbf{58.3} \\
Our (SAM) 5 shots  & \textbf{66.5 }& \textbf{63.6} & \textbf{66.8} & \textbf{63.1} \\
\hline
\end{tabular}
}
\end{center}
%\vspace{-0.5cm}
\end{table}

\begin{figure}
  \centering
  \resizebox{\columnwidth}{!}{
  \includegraphics[]{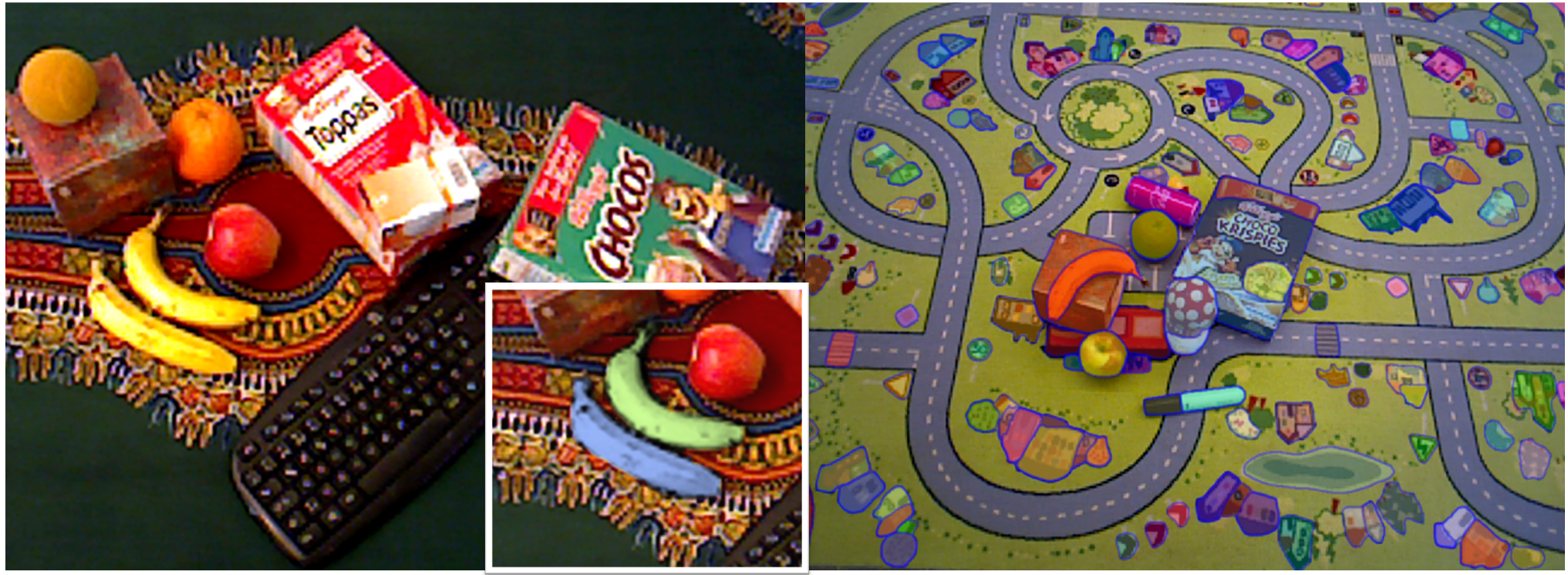}
  }
  \caption{The two characteristics of OCID that increased the difficulty in segmenting the target (left) the multiple instances of the same object (right) and the high variability of the background.}
  \label{fig:ocid_problems}
\end{figure}

\subsection{Performance assessment on Ocid}
\label{sec:ocid_res}

%To further test the generalization capabilities of the few-shot segmentation module in the few-shot semantic grasping, we also used the Ocid-grasp dataset. Table~\ref{tab:ocid} shows the results and the comparison with Grasping Siamese Mask R-CNN (GSMR-CNN)~\cite{holomjova2023gsmrGSMR} which, to the best of our knowledge, is the only work that uses Ocid-grasp for one-shot semantic grasping, similarly to what we propose in this work. 
%%%%%%%%%%%%%%% Riscaritta in caso di rimozione per non inziare con "Table"
The results on Ocid-grasp are reported in Table~\ref{tab:ocid},  comparing our approach with Grasping Siamese Mask R-CNN (GSMR-CNN)~\cite{holomjova2023gsmrGSMR} which, to the best of our knowledge, is the only work that uses Ocid-grasp for one-shot semantic grasping, similarly to what we propose in this work. 
%%%%%%%%%%%%%%%
From the comparison of the COCO metrics, the proposed approach shows better performance in recognizing the target objects---again, it should be pointed out that such results were obtained without any training or fine-tuning of the segmentator and the classifier on the Ocid dataset, differently from GSMR-CNN.

To compare with the results obtained on Graspnet-1B, we also computed mIoU, $A_{Sem}$ and $A_{SemGR}$. 
%%%%% PROVATO A RISCRIVERE
%The model shows good generalization capabilities to a different dataset, despite the drop in the mIoU, caused by two aspects visible in Figure~\ref{fig:ocid_problems}: (1) the variability of the background that makes the strategy of fitting the support plane less effective, and (2) the presence of multiple instances of the same class since our approach is capable of recognizing only one instance \ie the most suitable for grasping.
%%%%%
The model shows good generalization capabilities to a different dataset despite the presence of two hindering factors (Figure~\ref{fig:ocid_problems}): (1) the variability of the background that makes the strategy of fitting the support plane less effective, and (2) the presence of multiple instances of the same class since our approach is capable of recognizing only the instance most suitable for grasping. Both of these reduce the mIoU, since the first factor causes an increased number of mislead segmentations masks to input the few-shot classifier, whereas recognizing only one instance of a class impact the metric. This last factor is impacting the mIoU, but not the $A_{sem}$ that recovers from the decrease (mIoU 33.3\% and $A_{sem}$ 46.9\%) since the final objective is grasping a single object and not all the instances.

\begin{table}[t]
\caption{Few-shot segmentation results on OCID-grasp~\cite{ainetter2021endOCIDGRASP} in the object-wise split. AP, AP50 and AR1 are the COCO metrics, while mIoU is the mean Intersection over Union on the test objects. $A_{Sem}$ is the accuracy in selecting the correct target object, depicted by the shots. $A_{SemGR}$ considers also the grasp accuracy.}
\label{tab:ocid}
\begin{center}
\resizebox{\linewidth}{!}{
\begin{tabular}{|c|cccc|cc|}
\hline
& \multicolumn{4}{c|}{\rotatebox[origin=c]{0}{\color{black} Segmentation}} & \multicolumn{2}{c|}{\rotatebox[origin=c]{0}{\color{black} Grasp}}\\
\cline{2-7}
 \multirow{-2}{*}{ Model} & AP & AP50 & AR1 & mIoU & $A_{Sem}$ & $A_{SemGR}$  \\
\hline
GSMR-CNN~\cite{holomjova2023gsmrGSMR}    & 2.9 & 4.9 & 6.2 & - & - & - \\
%Our (SAM)      & \textbf{5.4} & \textbf{8.6} & \textbf{14.6} & 34.3 & 60.2 & 48.2  \\
Our (SAM) 1 shot     & \textbf{4.5} & \textbf{8.1} & \textbf{14.6} & 33.3 & 46.9 & 39.1  \\
\hline
\end{tabular}
}
\end{center}
\end{table}

\subsection{Real-world Experiments}

% Subtable 1
\begin{table}[t]
\caption{Real-world Experiments.}
\begin{subtable}{\columnwidth}
\centering
\begin{tabular}{|c|cc|cc|}
\hline
  & \multicolumn{2}{c|}{\rotatebox[origin=c]{0}{\color{black} 1-shot}} & \multicolumn{2}{c|}{\rotatebox[origin=c]{0}{\color{black} 5-shot}}\\
\cline{2-5}
 \multirow{-2}{*}{ Model} & $A_{Sem}$ & $A_{SemGR}$ & $A_{Sem}$ & $A_{SemGR}$ \\
\hline
FSG-Net~\cite{barcellona2023fsgFSGNET} & - & - & 38.3 & 32.3 \\
\hline
Our (SAM)  & 80.0 & 52.0 & \textbf{82.0 }& \textbf{54.0} \\
\hline
\end{tabular}
\vspace{5pt}
\caption{Results from Experiment A.}
\label{tab:realexp_sub_a}
\end{subtable}%

% Subtable 2
\begin{subtable}{\columnwidth}
\centering
\begin{tabular}{|c|cc|cc|}
\hline
  & \multicolumn{2}{c|}{\rotatebox[origin=c]{0}{\color{black} 1-shot}} & \multicolumn{2}{c|}{\rotatebox[origin=c]{0}{\color{black} 5-shot}}\\
\cline{2-5}
 \multirow{-2}{*}{Modality} & $A_{Sem}$ & $A_{SemGR}$ & $A_{Sem}$ & $A_{SemGR}$ \\
\hline
Instance-level & 95.6 & 56.5 & 100 & 65.2 \\
\hline
Class-level  & 48.1 & 40.7 & 64.8 & 48.1 \\
\hline
\end{tabular}
\vspace{5pt}
\caption{Results from Experiment B.}
\label{tab:realexp_sub_b}
\end{subtable}%

%\vspace{-0.5cm}
\end{table}

\begin{figure*}[ht!]
\centering
\resizebox{0.99\linewidth}{!}{
%\framebox{\parbox{3in}{Add image showing skeletons VS 3D Segmentation for qualitative comparison}}
\includegraphics[]{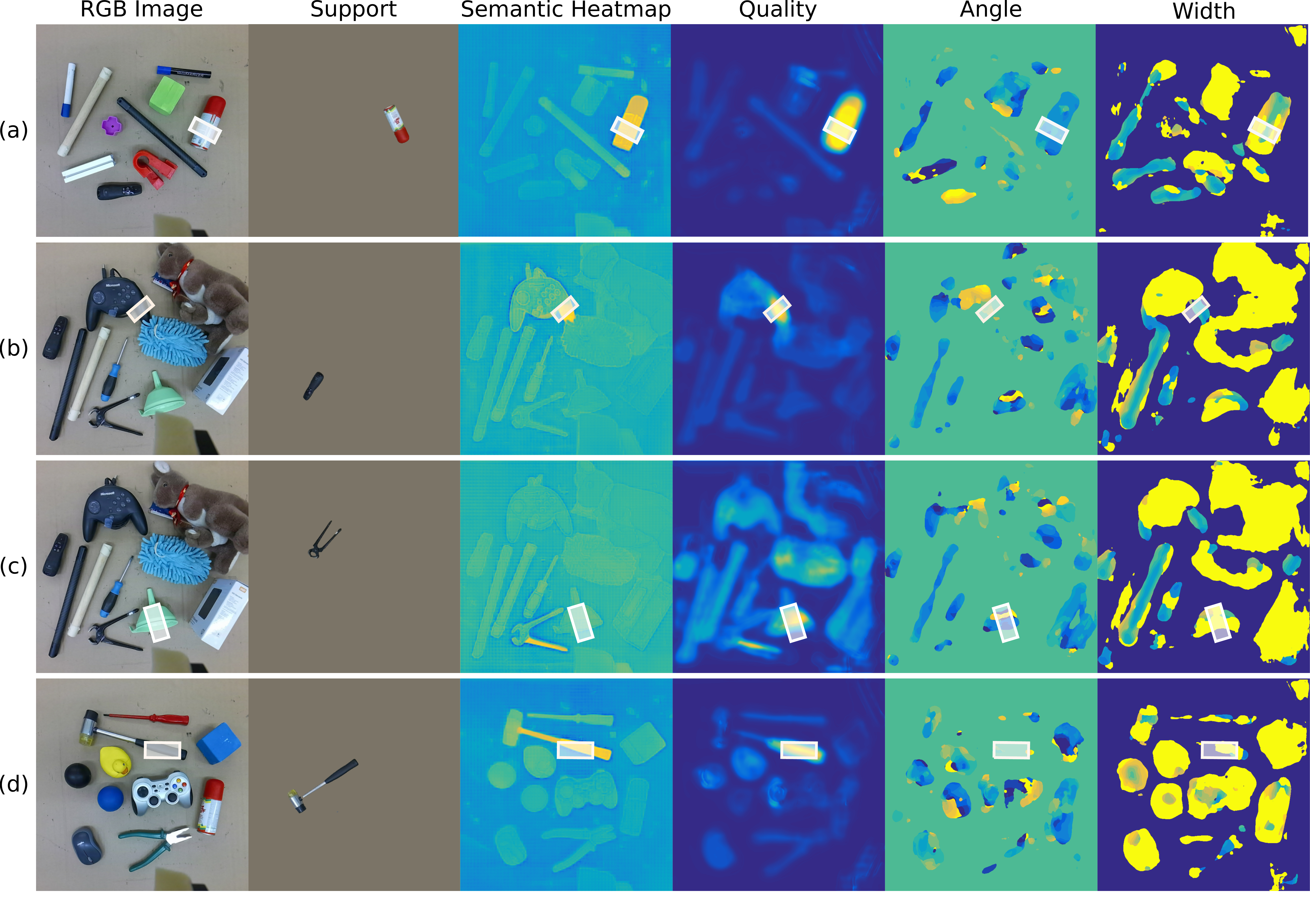}
}
\caption{The image shows a correct grasp prediction (a), a failure due to the few-shot module (b), a failure due to low confidence in the Quality module (c) and a failure due to a wrong grasp orientation estimation (d).}
%Colors distinguish different layer type, while numbers indicates the number of output channels from convolutional layers.
%(i.e. double Conv2D, Conv2D, MaxPool, Upsampling)
\label{fig:failure_modes}
\end{figure*}

%The evaluation in a real-world scenario with static cluttered objects confirmed the capabilities of generalization of the proposed method,  both in target recognition and grasping pose generation. 
The results of experiment A are reported in Table~\ref{tab:realexp_sub_a}. Compared with~\cite{barcellona2023fsgFSGNET}, where tests were conducted under similar real-world conditions, we observe significant improvements, particularly in the recognition of the target object. 

The semantic grasp generation shows a smaller gain, likely due to the increased number of objects taken into consideration.
Our real-world setup includes a larger set of objects with significant challenges in grasping such as pliers and screwdrivers. We identified 3 main source of failures of our model: (1) failures in the few-shot module, which leads to grasping a wrong object, particularly when objects have similar appearances, as shown in Figure~\ref{fig:failure_modes}b; (2) low confidence in the quality module predictions while fusing semantic heatmaps with the depth images may lead to the selection of a wrong target, as in Figure~\ref{fig:failure_modes}c; (3) angle or width wrong predictions mainly due to misalignment between the three heads which are trained separately%
%%%%%%% RISCRITTO
%, as shown in Figure~\ref{fig:failure_modes}d%
%%%% NUOVA VERSION
; the misalignment causes independent behaviours of the three heads that can produce prediction of an angle not compatible with the position and/or the width predicted, as visible in Figure~\ref{fig:failure_modes}d
%%%%%
. Other failures can be attributed to technical issues such as camera distortions in the peripheral area of the workspace and the physical gripper aperture limited to 10\,cm (this constraint on the aperture imposes restrictions on the range of feasible grasps).

The results from experiment B, as shown in Table~\ref{tab:realexp_sub_b}, clearly demonstrate the ability of the few-shot semantic segmentation module to recognize a specific instance among other objects of the same class (instance-level test).
%%%%% AGGIUNTO
Even with one example of the target, the model recognizes the correct instance among the same class 95.6\% of times.
%%%%%
%%%%%%%%%% RISCRITTO
%Even more noteworthy are the class-level tests, where the support set does not include the target object but a different instance of the same class (\ie semantic category). 
%%%%%%%%%%% NUOVA VERSIONE
Even more noteworthy are the class-level tests, where the support set includes a different instance of the object. Figure~\ref{fig:real_world_exp}B.2 shows an example of the experiment: the support image (right) contains examples of red pliers, while the target image (left) contains green pliers of a different type. 
%%%%%%%%%%%
%%%%%%%%%%% RISCRITTO
%The results are highly promising, in particular with 5 shots, indicating that the model can distinguish the correct semantic even when the visual appearance is different. 
%%%%%%%%%%% NUOVA VERSIONE
The results are remarkable: the model was able to correctly identify the target object 64.8\% of the times by using five examples of a different object, indicating that it can extract the correct semantic regardless of the different appearance. 
%%%%%%%%%%%
Deeper analysis of the results of the experiment allowed us to observe that the model tends to prioritize colour and simple shapes as discriminant features. Indeed, a perfect run was achieved using three balls of different color, thanks to the strong shape correlation. On the other hand, the proposed framework is challenged by markers and pliers, as they exhibit more varied appearances and shapes that can be easily confused with other objects.

\subsection{Limitations}
\label{sec:limit}
% spieghiamo i limiti dell'approccio che aprono la strada a nuovi lavori 
% 1 segmentazione produce troppi falsi positivi nel caso di ocid
% 2 prendiamo solo la maschera più probabile ignorando altre possibili istanze. 

%calcolo della z un po' alla buona --> passaggio a 6 DOF
%disallineamento tra le 3 head --> provoca valore 0 di width e angle anche nelle aree graspabili
%dare un support di immagini è sufficiente? 

As mentioned in section \ref{sec:ocid_res}, our approach shows some limitations. The main issue is the high false positive rate when the background is unstructured: with such scenes, SAM produces many possible targets, and the few-shot classifier struggles to predict the correct one. A possible solution would be fine-tuning SAM for the context, but this would come at the cost of a lower generalization capabilities.
%. However, it would also mean reducing the generalizability of the approach. 

The second main limitation is that the approach is not suitable for multiple instances. This issue is less impactful in the context of robot grasping because we can produce a correct pick by correctly detecting one instance. 
Nevertheless, it is critical in general few-shot semantic segmentation, where the goal is to predict the semantics for all the instances in the image.
This can be solved by improving the selection through the few-shot classifier: at the moment, only the most probable instance is chosen, whereas an absolute metric would guarantee a more mindful selection.

Regarding grasping pose synthesis,  we acknowledge that incorporating 7-DoF grasps \ie~$gr_{7} = (x, y, z, \theta, \phi, \gamma, w)$, which encompass the complete 6-DoF definition of the pose, would render the framework more suitable for manipulating larger and more complex objects. Moreover, we identified a limitation in the modular training approach proposed in~\cite{barcellona2023fsgFSGNET} during our experimentation: while this approach helps the training process by limiting the number of parameters updated at once and mitigating overfitting, it can lead to misalignments in the three heads; in particular, the Angle and Width modules tend to under-cover the actual graspable area.
Experiment B highlights an additional element that should be considered: is visual information enough? The usage of multi-modal input in Visual-Language Models (VLMs) has shown promising results in improving the model's generalization~\cite{radford2021learningCLIP, wei2023iclipICLIP}. The success of VLMs primarily stems from their large size, allowing them to encapsulate a vast knowledge base. However, this advantage comes at the cost of requiring large computational power, resulting in slower inference times and higher power consumption.
%%%%%%%%%%%% AGGIUNTO
This drawback is not negligible in applications requiring autonomous agents that need to plan their actions with limited time and hardware, such as autonomous robots. 
%%%%%%%%%%%%

% nonstante questo funziona bene
Despite these limitations, the approach demonstrated its effectiveness in robot grasping, and the range of application scenarios could become even wider when the current limitations will be overcome.

\section{CONCLUSIONS}
\label{sec:conclusions}
\label{sec:conclusions}
In this work, we presented a novel approach for few-shot semantic segmentation that leverages recent advances in foundation models for segmentation and few-shot classification. Foundation models were used as a zero-shot segmentator to predict a set of candidates, which were then discerned by the classifier. The latter was adapted to the task by inverting the support and query logic. Our novel approach demonstrated impressive results in the few-shot semantic segmentation, even without fine-tuning. When deployed for grasp synthesis, the proposed approach showed a significant improvement in selecting the correct object for semantic grasping. Exhaustive experiments in the real world revealed a small gap between the performance from the datasets to a real setup, highlighting the model's ability to detect the instance described by the support set even when it contains objects with different visual appearances. Importantly, in all experiments, the proposed few-shot segmentation module was not fine-tuned, meaning that the knowledge acquired during training can be easily exploited in several new scenarios.
%
%Despite these considerable advancements, several open questions persist. Can this approach be extended to multi-instance segmentation? Is it feasible to integrate segmentation into the 7-DoF grasp estimation? 
%Undoubtedly, the former would positively impact the research in computer vision, while the latter has the potential to create a significant impact in robotic grasping scenarios. 
%Moving towards 7-DoF grasps would also enable more complex manipulation tasks. For this reason, we will prioritize this research line for future investigations. 
The considerable advancements do not preclude further improvements. Multi-instance segmentation can be achieved by extracting an absolute metric from the classifier, while the few-shot segmentation heatmap can also benefit 7-DoF grasps, enabling more complex manipulation. Since achieving complex manipulation is essential for many tasks, we will prioritize this research line.
Another promising area is the exploitation of Visual-Language Models, which demonstrate impressive few-shot and zero-shot generalization capabilities across various tasks in the realm of robotics.


%%%%%%%%%%%%%%%%%%%%%%%%%%%%%%%%%%%%%%%%%%%%%%%%%%%%%%%%%%%%%%%%%%%%%%%%%%%%%%%%

%%%%%%%%%%%%%%%%%%%%%%%%%%%%%%%%%%%%%%%%%%%%%%%%%%%%%%%%%%%%%%%%%%%%%%%%%%%%%%%%

%%%%%%%%%%%%%%%%%%%%%%%%%%%%%%%%%%%%%%%%%%%%%%%%%%%%%%%%%%%%%%%%%%%%%%%%%%%%%%%%
%\section*{APPENDIX}

% \section*{ACKNOWLEDGMENT}
% Part of this work was supported by MIUR (Italian Minister for
% Education) under the initiative “PON Ricerca e Innovazione 2014 - 2020”, CUP C95F21007870007.

%%%%%%%%%%%%%%%%%%%%%%%%%%%%%%%%%%%%%%%%%%%%%%%%%%%%%%%%%%%%%%%%%%%%%%%%%%%%%%%%

%References are important to the reader; therefore, each citation must be complete and correct. If at all possible, references should be commonly available publications.

\bibliographystyle{IEEEtran}
\bibliography{newbibliography.bib}

\begin{thebibliography}{10}
\providecommand{\url}[1]{#1}
\csname url@rmstyle\endcsname
\providecommand{\newblock}{\relax}
\providecommand{\bibinfo}[2]{#2}
\providecommand\BIBentrySTDinterwordspacing{\spaceskip=0pt\relax}
\providecommand\BIBentryALTinterwordstretchfactor{4}
\providecommand\BIBentryALTinterwordspacing{\spaceskip=\fontdimen2\font plus
\BIBentryALTinterwordstretchfactor\fontdimen3\font minus \fontdimen4\font\relax}
\providecommand\BIBforeignlanguage[2]{{%
\expandafter\ifx\csname l@#1\endcsname\relax
\typeout{** WARNING: IEEEtran.bst: No hyphenation pattern has been}%
\typeout{** loaded for the language `#1'. Using the pattern for}%
\typeout{** the default language instead.}%
\else
\language=\csname l@#1\endcsname
\fi
#2}}

\bibitem{seita2021learningMANIPULATION}
D.~Seita, P.~Florence, J.~Tompson, E.~Coumans, V.~Sindhwani, K.~Goldberg, and A.~Zeng, ``Learning to rearrange deformable cables, fabrics, and bags with goal-conditioned transporter networks,'' in \emph{2021 IEEE International Conference on Robotics and Automation (ICRA)}.\hskip 1em plus 0.5em minus 0.4em\relax IEEE, 2021, pp. 4568--4575.

\bibitem{shridhar2023perceiverMANIPULATION2}
M.~Shridhar, L.~Manuelli, and D.~Fox, ``Perceiver-actor: A multi-task transformer for robotic manipulation,'' in \emph{Conference on Robot Learning}.\hskip 1em plus 0.5em minus 0.4em\relax PMLR, 2023, pp. 785--799.

\bibitem{geng2023rlaffordMANIPULATION3}
Y.~Geng, B.~An, H.~Geng, Y.~Chen, Y.~Yang, and H.~Dong, ``Rlafford: End-to-end affordance learning for robotic manipulation,'' in \emph{2023 IEEE International Conference on Robotics and Automation (ICRA)}.\hskip 1em plus 0.5em minus 0.4em\relax IEEE, 2023, pp. 5880--5886.

\bibitem{ggcnn}
\BIBentryALTinterwordspacing
D.~Morrison, P.~Corke, and J.~Leitner, ``Learning robust, real-time, reactive robotic grasping,'' \emph{The International Journal of Robotics Research}, vol.~39, no. 2-3, pp. 183--201, 2020. [Online]. Available: \url{https://doi.org/10.1177/0278364919859066}
\BIBentrySTDinterwordspacing

\bibitem{assembly}
L.~Zhou, L.~Zhang, and N.~Konz, ``Computer vision techniques in manufacturing,'' \emph{IEEE Transactions on Systems, Man, and Cybernetics: Systems}, vol.~53, no.~1, pp. 105--117, 2023.

\bibitem{sorting}
T.~Kiyokawa, J.~Takamatsu, and S.~Koyanaka, ``Challenges for future robotic sorters of mixed industrial waste: A survey,'' \emph{IEEE Transactions on Automation Science and Engineering}, vol.~21, no.~1, pp. 1023--1040, 2024.

\bibitem{christen2023learningHANDOVER}
S.~Christen, W.~Yang, C.~P{\'e}rez-D’Arpino, O.~Hilliges, D.~Fox, and Y.-W. Chao, ``Learning human-to-robot handovers from point clouds,'' in \emph{Proceedings of the IEEE/CVF Conference on Computer Vision and Pattern Recognition}, 2023, pp. 9654--9664.

\bibitem{duan2023semanticGRASPSEMANTIC}
S.~Duan, G.~Tian, Z.~Wang, S.~Liu, and C.~Feng, ``A semantic robotic grasping framework based on multi-task learning in stacking scenes,'' \emph{Engineering Applications of Artificial Intelligence}, vol. 121, p. 106059, 2023.

\bibitem{ainetter2021endOCIDGRASP}
S.~Ainetter and F.~Fraundorfer, ``End-to-end trainable deep neural network for robotic grasp detection and semantic segmentation from rgb,'' in \emph{2021 IEEE International Conference on Robotics and Automation (ICRA)}.\hskip 1em plus 0.5em minus 0.4em\relax IEEE, 2021, pp. 13\,452--13\,458.

\bibitem{liu2022unseenFSSRG}
X.~Liu, Y.~Zhang, and D.~Shan, ``Unseen object few-shot semantic segmentation for robotic grasping,'' \emph{IEEE Robotics and Automation Letters}, vol.~8, no.~1, pp. 320--327, 2022.

\bibitem{holomjova2023gsmrGSMR}
V.~Holomjova, A.~J. Starkey, and P.~Mei{\ss}ner, ``Gsmr-cnn: An end-to-end trainable architecture for grasping target objects from multi-object scenes,'' in \emph{2023 IEEE International Conference on Robotics and Automation (ICRA)}.\hskip 1em plus 0.5em minus 0.4em\relax IEEE, 2023, pp. 3808--3814.

\bibitem{barcellona2023fsgFSGNET}
L.~Barcellona, A.~Bacchin, A.~Gottardi, E.~Menegatti, and S.~Ghidoni, ``Fsg-net: a deep learning model for semantic robot grasping through few-shot learning,'' in \emph{2023 IEEE International Conference on Robotics and Automation (ICRA)}.\hskip 1em plus 0.5em minus 0.4em\relax IEEE, 2023, pp. 1793--1799.

\bibitem{boudiaf2021fewNOMETA}
M.~Boudiaf, H.~Kervadec, Z.~I. Masud, P.~Piantanida, I.~Ben~Ayed, and J.~Dolz, ``Few-shot segmentation without meta-learning: A good transductive inference is all you need?'' in \emph{Proceedings of the IEEE/CVF conference on computer vision and pattern recognition}, 2021, pp. 13\,979--13\,988.

\bibitem{hu2022pushingPMF}
S.~X. Hu, D.~Li, J.~St{\"u}hmer, M.~Kim, and T.~M. Hospedales, ``Pushing the limits of simple pipelines for few-shot learning: External data and fine-tuning make a difference,'' in \emph{Proceedings of the IEEE/CVF Conference on Computer Vision and Pattern Recognition}, 2022, pp. 9068--9077.

\bibitem{dosovitskiy2020imageVIT}
A.~Dosovitskiy, L.~Beyer, A.~Kolesnikov, D.~Weissenborn, X.~Zhai, T.~Unterthiner, M.~Dehghani, M.~Minderer, G.~Heigold, S.~Gelly, \emph{et~al.}, ``An image is worth 16x16 words: Transformers for image recognition at scale,'' \emph{arXiv preprint arXiv:2010.11929}, 2020.

\bibitem{wei2023iclipICLIP}
Y.~Wei, Y.~Cao, Z.~Zhang, H.~Peng, Z.~Yao, Z.~Xie, H.~Hu, and B.~Guo, ``iclip: Bridging image classification and contrastive language-image pre-training for visual recognition,'' in \emph{Proceedings of the IEEE/CVF Conference on Computer Vision and Pattern Recognition}, 2023, pp. 2776--2786.

\bibitem{kirillov2023segmentSAM}
A.~Kirillov, E.~Mintun, N.~Ravi, H.~Mao, C.~Rolland, L.~Gustafson, T.~Xiao, S.~Whitehead, A.~C. Berg, W.-Y. Lo, \emph{et~al.}, ``Segment anything,'' \emph{arXiv preprint arXiv:2304.02643}, 2023.

\bibitem{wang2023cutCUTLER}
X.~Wang, R.~Girdhar, S.~X. Yu, and I.~Misra, ``Cut and learn for unsupervised object detection and instance segmentation,'' in \emph{Proceedings of the IEEE/CVF Conference on Computer Vision and Pattern Recognition}, 2023, pp. 3124--3134.

\bibitem{ye2023gaussian3DRECONSTRUCTION}
M.~Ye, M.~Danelljan, F.~Yu, and L.~Ke, ``Gaussian grouping: Segment and edit anything in 3d scenes,'' \emph{arXiv preprint arXiv:2312.00732}, 2023.

\bibitem{qin2023langsplatRECONSTRUCTIONLANGUAGE}
M.~Qin, W.~Li, J.~Zhou, H.~Wang, and H.~Pfister, ``Langsplat: 3d language gaussian splatting,'' \emph{arXiv preprint arXiv:2312.16084}, 2023.

\bibitem{fang2020graspnetGRASPNET}
H.-S. Fang, C.~Wang, M.~Gou, and C.~Lu, ``Graspnet-1billion: A large-scale benchmark for general object grasping,'' in \emph{Proceedings of the IEEE/CVF conference on computer vision and pattern recognition}, 2020, pp. 11\,444--11\,453.

\bibitem{antipodal}
S.~Kumra, S.~Joshi, and F.~Sahin, ``Antipodal robotic grasping using generative residual convolutional neural network,'' in \emph{2020 IEEE/RSJ International Conference on Intelligent Robots and Systems (IROS)}, 2020, pp. 9626--9633.

\bibitem{li2021adaptiveASGNET}
G.~Li, V.~Jampani, L.~Sevilla-Lara, D.~Sun, J.~Kim, and J.~Kim, ``Adaptive prototype learning and allocation for few-shot segmentation,'' in \emph{Proceedings of the IEEE/CVF conference on computer vision and pattern recognition}, 2021, pp. 8334--8343.

\bibitem{shaban2017oneFSS2017}
A.~Shaban, S.~Bansal, Z.~Liu, I.~Essa, and B.~Boots, ``One-shot learning for semantic segmentation,'' \emph{arXiv preprint arXiv:1709.03410}, 2017.

\bibitem{tian2020priorPFENET}
Z.~Tian, H.~Zhao, M.~Shu, Z.~Yang, R.~Li, and J.~Jia, ``Prior guided feature enrichment network for few-shot segmentation,'' \emph{IEEE transactions on pattern analysis and machine intelligence}, vol.~44, no.~2, pp. 1050--1065, 2020.

\bibitem{zhang2021selfCANET}
B.~Zhang, J.~Xiao, and T.~Qin, ``Self-guided and cross-guided learning for few-shot segmentation,'' in \emph{Proceedings of the IEEE/CVF Conference on Computer Vision and Pattern Recognition}, 2021, pp. 8312--8321.

\bibitem{dong2018fewPROTO}
N.~Dong and E.~P. Xing, ``Few-shot semantic segmentation with prototype learning.'' in \emph{BMVC}, vol.~3, no.~4, 2018.

\bibitem{wang2019panetPANET}
K.~Wang, J.~H. Liew, Y.~Zou, D.~Zhou, and J.~Feng, ``Panet: Few-shot image semantic segmentation with prototype alignment,'' in \emph{proceedings of the IEEE/CVF international conference on computer vision}, 2019, pp. 9197--9206.

\bibitem{sun2022singularBAM}
Y.~Sun, Q.~Chen, X.~He, J.~Wang, H.~Feng, J.~Han, E.~Ding, J.~Cheng, Z.~Li, and J.~Wang, ``Singular value fine-tuning: Few-shot segmentation requires few-parameters fine-tuning,'' \emph{Advances in Neural Information Processing Systems}, vol.~35, pp. 37\,484--37\,496, 2022.

\bibitem{peng2023hierarchicalHDMNET}
B.~Peng, Z.~Tian, X.~Wu, C.~Wang, S.~Liu, J.~Su, and J.~Jia, ``Hierarchical dense correlation distillation for few-shot segmentation,'' in \emph{Proceedings of the IEEE/CVF Conference on Computer Vision and Pattern Recognition}, 2023, pp. 23\,641--23\,651.

\bibitem{zhao2023surveyLLMSURVEY}
W.~X. Zhao, K.~Zhou, J.~Li, T.~Tang, X.~Wang, Y.~Hou, Y.~Min, B.~Zhang, J.~Zhang, Z.~Dong, \emph{et~al.}, ``A survey of large language models,'' \emph{arXiv preprint arXiv:2303.18223}, 2023.

\bibitem{saviolo2023unifyingSAMTRACKING}
A.~Saviolo, P.~Rao, V.~Radhakrishnan, J.~Xiao, and G.~Loianno, ``Unifying foundation models with quadrotor control for visual tracking beyond object categories,'' \emph{arXiv preprint arXiv:2310.04781}, 2023.

\bibitem{nanni2023improvingSAMZS}
L.~Nanni, D.~Fusaro, C.~Fantozzi, and A.~Pretto, ``Improving existing segmentators performance with zero-shot segmentators,'' \emph{Entropy}, vol.~25, no.~11, p. 1502, 2023.

\bibitem{caron2021emergingDINO}
M.~Caron, H.~Touvron, I.~Misra, H.~J{\'e}gou, J.~Mairal, P.~Bojanowski, and A.~Joulin, ``Emerging properties in self-supervised vision transformers,'' in \emph{Proceedings of the IEEE/CVF international conference on computer vision}, 2021, pp. 9650--9660.

\bibitem{vaswani2017attentionTRANSFORMER}
A.~Vaswani, N.~Shazeer, N.~Parmar, J.~Uszkoreit, L.~Jones, A.~N. Gomez, {\L}.~Kaiser, and I.~Polosukhin, ``Attention is all you need,'' \emph{Advances in neural information processing systems}, vol.~30, 2017.

\bibitem{snell2017prototypicalPROTONET}
J.~Snell, K.~Swersky, and R.~Zemel, ``Prototypical networks for few-shot learning,'' \emph{Advances in neural information processing systems}, vol.~30, 2017.

\bibitem{okazawa2022interclass1way1}
A.~Okazawa, ``Interclass prototype relation for few-shot segmentation,'' in \emph{European Conference on Computer Vision}.\hskip 1em plus 0.5em minus 0.4em\relax Springer, 2022, pp. 362--378.

\bibitem{sung2018learningFSCMETRIC}
F.~Sung, Y.~Yang, L.~Zhang, T.~Xiang, P.~H. Torr, and T.~M. Hospedales, ``Learning to compare: Relation network for few-shot learning,'' in \emph{Proceedings of the IEEE conference on computer vision and pattern recognition}, 2018, pp. 1199--1208.

\bibitem{wertheimer2021fewRECONSTRUCTION}
D.~Wertheimer, L.~Tang, and B.~Hariharan, ``Few-shot classification with feature map reconstruction networks,'' in \emph{Proceedings of the IEEE/CVF conference on computer vision and pattern recognition}, 2021, pp. 8012--8021.

\bibitem{hiller2022rethinkingFEWTURE}
M.~Hiller, R.~Ma, M.~Harandi, and T.~Drummond, ``Rethinking generalization in few-shot classification,'' \emph{Advances in Neural Information Processing Systems}, vol.~35, pp. 3582--3595, 2022.

\bibitem{he2022maskedMASKED}
K.~He, X.~Chen, S.~Xie, Y.~Li, P.~Doll{\'a}r, and R.~Girshick, ``Masked autoencoders are scalable vision learners,'' in \emph{Proceedings of the IEEE/CVF conference on computer vision and pattern recognition}, 2022, pp. 16\,000--16\,009.

\bibitem{tancik2020fourierPOSITIONALENCODING}
M.~Tancik, P.~Srinivasan, B.~Mildenhall, S.~Fridovich-Keil, N.~Raghavan, U.~Singhal, R.~Ramamoorthi, J.~Barron, and R.~Ng, ``Fourier features let networks learn high frequency functions in low dimensional domains,'' \emph{Advances in Neural Information Processing Systems}, vol.~33, pp. 7537--7547, 2020.

\bibitem{unet}
O.~Ronneberger, P.~Fischer, and T.~Brox, ``U-net: Convolutional networks for biomedical image segmentation,'' in \emph{Medical Image Computing and Computer-Assisted Intervention -- MICCAI 2015}, N.~Navab, J.~Hornegger, W.~M. Wells, and A.~F. Frangi, Eds.\hskip 1em plus 0.5em minus 0.4em\relax Springer International Publishing, 2015, pp. 234--241.

\bibitem{suchi2019easylabelOCID}
M.~Suchi, T.~Patten, D.~Fischinger, and M.~Vincze, ``Easylabel: A semi-automatic pixel-wise object annotation tool for creating robotic rgb-d datasets,'' in \emph{2019 International Conference on Robotics and Automation (ICRA)}.\hskip 1em plus 0.5em minus 0.4em\relax IEEE, 2019, pp. 6678--6684.

\bibitem{loghmani2018recognizingARID}
M.~R. Loghmani, B.~Caputo, and M.~Vincze, ``Recognizing objects in-the-wild: Where do we stand?'' in \emph{2018 IEEE International Conference on Robotics and Automation (ICRA)}.\hskip 1em plus 0.5em minus 0.4em\relax IEEE, 2018, pp. 2170--2177.

\bibitem{calli2017yaleYCB}
B.~Calli, A.~Singh, J.~Bruce, A.~Walsman, K.~Konolige, S.~Srinivasa, P.~Abbeel, and A.~M. Dollar, ``Yale-cmu-berkeley dataset for robotic manipulation research,'' \emph{The International Journal of Robotics Research}, vol.~36, no.~3, pp. 261--268, 2017.

\bibitem{evangelista2022unifiedCALIBRATION}
D.~Evangelista, D.~Allegro, M.~Terreran, A.~Pretto, and S.~Ghidoni, ``An unified iterative hand-eye calibration method for eye-on-base and eye-in-hand setups,'' in \emph{2022 IEEE 27th International Conference on Emerging Technologies and Factory Automation (ETFA)}.\hskip 1em plus 0.5em minus 0.4em\relax IEEE, 2022, pp. 1--7.

\bibitem{lu2021simplerCWT}
Z.~Lu, S.~He, X.~Zhu, L.~Zhang, Y.-Z. Song, and T.~Xiang, ``Simpler is better: Few-shot semantic segmentation with classifier weight transformer,'' in \emph{Proceedings of the IEEE/CVF International Conference on Computer Vision}, 2021, pp. 8741--8750.

\bibitem{radford2021learningCLIP}
A.~Radford, J.~W. Kim, C.~Hallacy, A.~Ramesh, G.~Goh, S.~Agarwal, G.~Sastry, A.~Askell, P.~Mishkin, J.~Clark, \emph{et~al.}, ``Learning transferable visual models from natural language supervision,'' in \emph{International conference on machine learning}.\hskip 1em plus 0.5em minus 0.4em\relax PMLR, 2021, pp. 8748--8763.

\end{thebibliography}

\end{document}